\documentclass[sigconf]{acmart}

\usepackage{microtype}
\usepackage{graphicx}
\usepackage{subfigure}
\usepackage{booktabs} 
\usepackage{soul}
\usepackage{color}
\usepackage{graphicx}
\usepackage{url}
\usepackage{amsmath}
\usepackage{amsthm}
\usepackage{amsfonts}
\usepackage{algorithm}
\usepackage{algorithmic}

\usepackage{xcolor}
\usepackage[normalem]{ulem}
\usepackage{multirow}
\usepackage{multicol}
\usepackage{tikz}
\usepackage{pgfplots}
\usepackage{colortbl}
\usepackage{array}
\usepackage{pifont}
\usepackage{listings}

\newcommand*\circled[1]{\tikz[baseline=(char.base)]{
            \node[shape=circle,draw,inner sep=0.4pt] (char) {#1};}}
\renewcommand\footnotetextcopyrightpermission[1]{}
\acmConference[]{}{}{}
\begin{document}

\title{Fed-pilot: Optimizing LoRA Allocation for Efficient Federated Fine-Tuning with Heterogeneous Clients}

\author{Zikai Zhang, Rui Hu, Ping Liu, Jiahao Xu}
\email{{zikaiz, ruihu, pingl, jiahaox}@unr.edu}
\affiliation{%
  \institution{University of Nevada, Reno}
  \city{Reno}
  \state{NV}
  \country{USA}
}

\renewcommand{\shortauthors}{Zhang et al.}

\begin{abstract}
Federated Learning enables the fine-tuning of foundation models (FMs) across distributed clients for specific tasks; however, its scalability is limited by the heterogeneity of client memory capacities. In this work, we propose Fed-pilot, a memory-efficient federated fine-tuning framework. It enables memory-constrained clients to participate in Low-Rank Adaptation (LoRA)-based fine-tuning by training only a subset of LoRA modules locally.
Fed-pilot identifies the optimal selection of trainable LoRA modules as a knapsack optimization problem, maximizing model performance under memory constraints for each client. To mitigate inconsistencies arising from heterogeneous module allocations and Non-IID data, Fed-pilot employs a novel aggregation rule that dynamically compensates for under-updated layers. Extensive experiments on five diverse datasets across various heterogeneous data settings demonstrate Fed-pilot's effectiveness and efficiency compared to state-of-the-art methods. To the best of our knowledge, this is the first study on federated fine-tuning of FMs that integrates memory-constrained optimization. The code will be publicly available.
\end{abstract}




\maketitle

\section{Introduction}
\label{sec:introduction}

\begin{figure}[t]
\centering
\includegraphics[width=0.48 \textwidth]{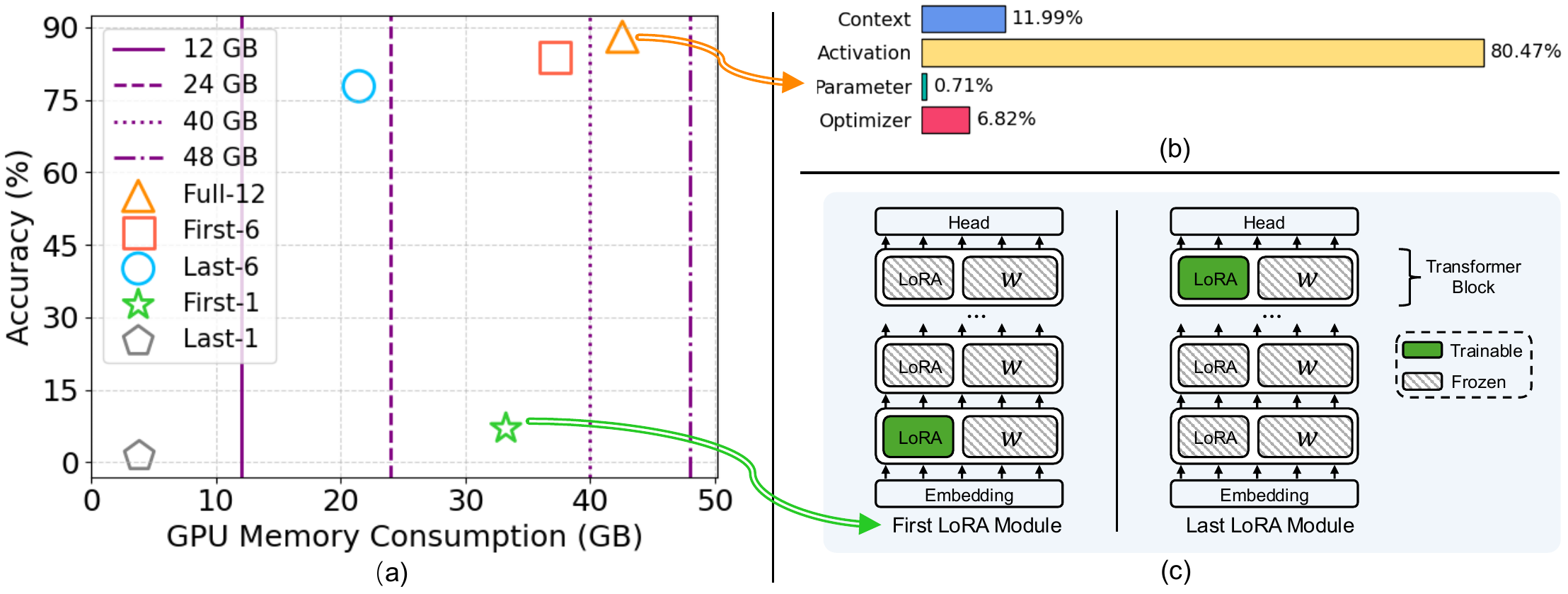}
\caption{(a) Accuracy vs. GPU memory consumption for fine-tuning the ViT-base model with trainable LoRA modules at different positions. (b) Breakdown of memory consumption of LoRA fine-tuning. (c) Comparison of training the first LoRA module vs. the last LoRA module while keeping the rest frozen. See {Supplement~\ref{subsec:appendix-motivation-settings}} for experimental details.}
\label{fig:motivation}
\end{figure}

Federated Learning (FL)~\cite{mcmahan2017communication} enables collaborative model training across distributed edge devices (clients) under the coordination of a central server. In this paradigm, data remains local, and only client-computed model updates are sent to the server to iteratively refine a shared global model. By reducing data collection costs and mitigating privacy risks, FL has emerged as a practical solution for scalable machine learning systems. 
%
Concurrently, Foundation Models (FMs) have become transformative across various domains due to their large scale and ability to generalize effectively across diverse datasets. These models have driven significant advancements in natural language processing, computer vision, and other fields through task-specific fine-tuning~\cite{touvron2023llama,dosovitskiy2020image}. Recent developments, such as the Llama-family models~\cite{touvron2023llama}, exemplify the rapid expansion of FMs to billions of parameters, showcasing their unprecedented scale and complexity. 
However, the large parameter count of FMs poses significant challenges for deployment in federated settings, as the associated computational and communication overhead severely limits their ability to efficiently utilize distributed data~\cite{su2023fedra}. 


To address the challenges of deploying FMs in federated settings, Parameter-Efficient Fine-Tuning (PEFT)~\cite{lester2021power} methods have been widely explored. Among these, Low-Rank Adaptation (LoRA) has demonstrated strong performance in efficiently fine-tuning FMs, particularly on mid-sized datasets~\cite{hu2021lora}. 
LoRA achieves this by decomposing weight updates into low-rank matrices, training only these matrices while keeping the pre-trained weights frozen. This approach significantly reduces the number of trainable parameters, lowering both the computational cost of training and the communication overhead for transmitting model updates in FL. 
%
%
%
Recent studies have explored LoRA's effectiveness in federated fine-tuning~\cite{gao2025flowertune,zhang2025fed,zhang2023towards,cho2024heterogeneous,wangflora,bai2024federated}.
For example, FedIT~\cite{zhang2023towards} proposes a vanilla integration of LoRA and FedAvg~\cite{mcmahan2017communication}, while methods like
HETLoRA~\cite{cho2024heterogeneous}, FLoRA~\cite{wangflora}, and FlexLoRA~\cite{bai2024federated} adjust LoRA ranks to accommodate clients' varying computational capabilities, optimizing resource usage in heterogeneous settings.

However, these methods fail to address a critical challenge in fine-tuning FMs in federated settings: the memory bottleneck. Fine-tuning FMs requires substantial on-device GPU memory to store the CUDA context, model parameters, optimizer states, and intermediate activations necessary for gradient computation during back-propagation. Among these, activations dominate memory usage, as illustrated in Figure~\ref{fig:motivation}(b). 
%
Although reducing the rank of LoRA modules lowers memory costs associated with optimizer states (e.g., gradients, momentum), it does not significantly alleviate the memory demand for activations.
This memory requirement makes fine-tuning FMs challenging even on powerful GPU servers, let alone on those resource-constrained edge or end devices in federated settings. 
%

Recently, FedRA~\cite{su2023fedra} was proposed to mitigate the memory bottleneck by training only a random subset of LoRA modules on each client, while keeping the remaining modules frozen. However, this approach assumes that freezing any LoRA module results in equal memory savings and overlooks the varying importance of different modules in contributing to model performance, thereby resulting in limited memory efficiency. 
As shown in Figure~\ref{fig:motivation}(a), training the LoRA modules
at different positions (i.e., within different transformer blocks) leads to varying memory consumption and model accuracy. This discrepancy arises because freezing a layer eliminates the storage of dynamic activations, which are associated with linear functions, but does not reduce the memory required for static activations. Static activations, which are generated by non-linear functions, must be stored for gradient computation in subsequent trainable layers (refer to {{Supplement~\ref{sec:appendix-gpu-memory-consumption}}} for details). For example, training the first $u$ LoRA modules requires extra storage for static activations from module $u+1$ onward to compute gradients back to the $u$-th module, even though these modules are frozen.

In this work, we introduce \textbf{Fed-pilot}, a novel memory-efficient method for \textbf{\uline{fed}}erated fine-tuning of FMs across clients with heterogeneous memory capacities. Fed-pilot identifies the o\textbf{\uline{p}}t\textbf{\uline{i}}mal al\textbf{\uline{lo}}ca\textbf{\uline{t}}ion of local trainable LoRA modules that maximize global model performance under memory constraints. To aggregate local LoRA updates, Fed-pilot introduces a compensated aggregation rule, which mitigates the inconsistencies across layers caused by Non-IID data distributions and heterogeneous LoRA allocation to ensure model convergence.
%
%
%
%
%
Our main contributions are summarized as follows:

\begin{itemize}
    \item We propose Fed-pilot, a novel federated LoRA-based fine-tuning framework designed for clients with heterogeneous memory constraints. Fed-pilot enables clients with limited memory to dynamically train an optimal subset of LoRA modules, which can effectively reduce memory requirements during fine-tuning. 
    To the best of our knowledge, this is the first study on federated fine-tuning of FMs that incorporates memory-constrained optimization. 
    \item We formulate the allocation of trainable LoRA modules as a knapsack optimization problem, aiming to maximize the fine-tuning performance under memory constraints. We model the GPU memory consumption of LoRA-based fine-tuning, including scenarios with partial LoRA allocation. We design an information gain (IG) score based value function to quantify the LoRA module's contribution to model performance from local and global perspectives. 
    
    \item In addition, we design a novel aggregation rule ComAgg to mitigate the significant divergence among local model updates caused by heterogeneous LoRA allocation and Non-IID data.
    \item We conduct extensive experiments using five datasets in IID and various Non-IID settings and compare Fed-pilot with state-of-the-art methods to demonstrate its effectiveness and efficiency.
\end{itemize}


\section{Related Work}\label{relatedwork}

\textbf{LoRA-based Federated Fine-Tuning.} FL has been applied to the PEFT of FMs, enabling multiple clients to collaboratively fine-tune FMs without sharing their data~\cite{zhang2023towards,babakniya2023slora,sun2024improving}.
There is growing interest in LoRA-based federated fine-tuning approaches, which are commonly known to be particularly effective in mid-sized data scenarios~\cite{hu2021lora}. 
FedIT~\cite{zhang2023towards} proposes a vanilla federated fine-tuning method with LoRA and establishes benchmarks for PEFT in FL. SLoRA~\cite{babakniya2023slora} is a two-stage approach that trains local models with sparse fine-tuning for several rounds and then switches to LoRA for PEFT. FFA-LoRA~\cite{sun2024improving} enhances the privacy of federated fine-tuning using differential privacy. {FedSA-LoRA}~\cite{guo2024selective} only shares the A matrices with the server for aggregation, due to their specific role in learning general knowledge. However, these methods assume homogeneous client resources.

\textbf{Federated Fine-Tuning under Heterogeneous Clients.} While LoRA-based federated fine-tuning methods can save computational and communication costs, they still fail to fully accommodate training across heterogeneous clients with varying memory capacities. A range of rank-based methods~\cite{cho2024heterogeneous,wangflora,bai2024federated} attempt to address this by assigning different LoRA ranks according to clients' memory capacities.
HETLoRA~\cite{cho2024heterogeneous} introduces a heterogeneous LoRA rank strategy, where clients train different numbers of LoRA ranks based on their memory capacity. Its aggregation rule follows a hierarchical approach, ranks shared by multiple clients are averaged, while higher ranks are aggregated by fewer clients.
FLoRA~\cite{wangflora} introduces a stacking method for all LoRA ranks to mitigate the aggregation noise from aggregating low-rank compositions separately. FlexLoRA~\cite{bai2024federated} utilizes singular value decomposition (SVD) to redistribute attention weight matrix for clients with varying ranks. However, reducing the LoRA rank does not result in significant memory reduction, making it an inefficient solution for memory-constrained settings.
FedRA~\cite{su2023fedra} proposes a layer-wise LoRA allocation strategy that randomly selects partial LoRA modules each round to handle resource disparities among clients. However, it fails to account for the non-linear nature of memory consumption and the varying impact of different layers on overall global model performance.
A well-designed heterogeneous LoRA allocation is essential for achieving optimal performance in federated fine-tuning systems with memory constraints. 

\textbf{Layer Freezing for Training Efficiency.} 
{
Layer freezing has been used to improve the training efficiency of FMs}~\cite{pan2024lisa,zhu2023lift,yao2024layer}.
LISA~\cite{pan2024lisa} randomly selects a subset of layers to fine-tune in each training step, thereby reducing the number of trainable parameters. LIFT~\cite{zhu2023lift} fine-tunes only one FM layer per step and applies different selection strategies throughout training, leading to reduced computational costs. IST~\cite{yao2024layer} introduces a reinforcement learning-based method to select a subset of layers during fine-tuning. Other methods in FL~\cite{liu2024fisher,wu2024fedfmsl} also incorporate layer freezing to improve training efficiency. 
FibecFed~\cite{liu2024fisher} designed a Fisher Information Matrix-based scoring mechanism to selectively fine-tune a subset of LoRA modules for each client.
FedFMSL~\cite{wu2024fedfmsl} progressively unfreezes LoRA modules from higher to shallow layers for efficient training, relying on heuristic training-time accuracy gains to determine when to unfreeze more LoRA modules.
However, these methods do not account for practical memory modeling, particularly the non-linear activation memory reduction by layer freezing. Our method optimizes LoRA allocation under memory constraints while preserving the global model performance.

\section{System Modeling}\label{sec:sys_model}


\subsection{Federated Fine-Tuning System}\label{sec:leaning_model}
We consider a federated fine-tuning system consisting of a central server and $v$ clients. The server has a global FM $\boldsymbol{\Theta}$ initialized with pre-trained weights $\boldsymbol{\Theta}^0$. Each client $i\in[v]$ owns a local dataset $D_i\sim\mathcal{P}_i$, where the size of $D_i$ and the data distribution $\mathcal{P}_i$ can vary across clients. In federated fine-tuning, clients aim to collaboratively fine-tune the global model $\boldsymbol{\Theta}$ under the coordination of the server.

In this work, we focus on transformer-based models~\cite{touvron2023llama,zhang2022opt,devlin2019bert,dosovitskiy2020image} consisting of $l$ transformer blocks $\{\boldsymbol{\mathcal{W}}_{(1)}, \cdots, \boldsymbol{\mathcal{W}}_{(l)}\}$, and embedding layers, along with head layers, denoted as $\boldsymbol{\mathcal{W}}_{rest}$. The pre-trained model is thus represented as $\boldsymbol{\Theta}^0=\{\boldsymbol{\mathcal{W}}_{(1)}^0, \cdots, \boldsymbol{\mathcal{W}}_{(l)}^0, \boldsymbol{\mathcal{W}}_{rest}^0\}$. 
To fine-tune the global model efficiently, we apply the LoRA module (denoted as $\boldsymbol{\theta}_{(j)}$) to the attention weight $\boldsymbol{\mathcal{A}}_{(j)}\in\mathbb{R}^{d \times k}$ within the $j$-th transformer block, where $j\in[1,l]$, and $d$ and $k$ represent the dimensions of the input and output feature space of attention layers, respectively. 
The idea behind LoRA is to constrain $\Delta\boldsymbol{\mathcal{A}}_{(j)}$, the update of a pre-trained attention weight matrix $\boldsymbol{\mathcal{A}}_{(j)}^0$, using a low-rank decomposition, such that $\boldsymbol{\mathcal{A}}_{(j)}^0 + \Delta\boldsymbol{\mathcal{A}}_{(j)}=\boldsymbol{\mathcal{A}}_{(j)}^0+\boldsymbol{\mathcal{N}}_{(j)}\boldsymbol{\mathcal{M}}_{(j)}$, where $\boldsymbol{\mathcal{N}}_{(j)}\in\mathbb{R}^{d \times r}$, $\boldsymbol{\mathcal{M}}_{(j)}\in\mathbb{R}^{r \times k}$, and the rank $r \ll \min(d, k)$. Without loss of generality for subsequent optimization, we omit $\boldsymbol{\mathcal{W}}_{rest}$ and define the LoRA modules for $\boldsymbol{\Theta}$ 
as $\boldsymbol{\theta} := \{ \boldsymbol{\theta}_{(j)}\}_{j\in[l]}$ with $\boldsymbol{\theta}_{(j)} = \{\boldsymbol{\mathcal{N}}_{(j)},\boldsymbol{\mathcal{M}}_{(j)}\}$.

%
In this context, the problem of federated fine-tuning using LoRA can be formulated as follows:
\begin{equation}\label{eq:ffm_obj}
\min_{\boldsymbol{\mathcal{\theta}}} \mathcal{L}(\boldsymbol{\mathcal{\theta}}) := \frac{1}{v}\sum_{i=1}^{v} \mathcal{L}_{i}(\boldsymbol{\mathcal{\theta}}; \boldsymbol{\Theta}^0),
\end{equation}
where $\mathcal{L}_{i}(\boldsymbol{\mathcal{\theta}}; \boldsymbol{\Theta}^0) := \mathbb{E}_{x \in  D_i}[\ell(\boldsymbol{\mathcal{\theta}};\boldsymbol{\Theta}^0,x)]$ denotes the local objective function of client $i$, and $\ell(\boldsymbol{\mathcal{\theta}}_i;\boldsymbol{\Theta}^0,x)$ is the loss of the model $\{\boldsymbol{\mathcal{\theta}}, \boldsymbol{\Theta}^0 \}$ on a datapoint $x$ sampled from ${D}_i$. 
We assume that clients download the pre-trained weights $\boldsymbol{\Theta}^0$ from the server prior to fine-tuning and have sufficient memory capacity to store the model parameters. Ideally, the classic FedAvg algorithm~\cite{mcmahan2017communication} can be employed to solve this problem in a distributed manner. In FedAvg, at round $t$, clients download the global trainable model parameters $\boldsymbol{\mathcal{\theta}}^t$ maintained by the server and update it using their local datasets to minimize their local objectives. Clients then upload their model updates to the server, which aggregates the updates to compute $\boldsymbol{\mathcal{\theta}}^{t+1}$ for the next round of fine-tuning. 
However, in practice, the large memory requirements for fine-tuning FMs, coupled with the heterogeneity of clients' memory capacities, prevent some clients from accommodating the memory cost necessary for locally optimizing $\boldsymbol{\mathcal{\theta}}$. 

\subsection{On-Device GPU Memory Consumption of LoRA-based Fine-Tuning}\label{sec:memory_consumption}
FM fine-tuning requires substantial on-device GPU memory, primarily due to the complexity of the transformer model architecture. In federated settings with memory-limited clients, understanding the memory footprint of a FM fine-tuning task is crucial for designing memory-efficient approaches. 
%
%
Let $\mathcal{C}$ denotes the total memory consumption for a LoRA-based FM fine-tuning process. It can be represented as $\mathcal{C}=\mathcal{C}_p+\mathcal{C}_o+\mathcal{C}_a+\mathcal{C}_c$, where $\mathcal{C}_p$, $\mathcal{C}_o$, $\mathcal{C}_a$ and $\mathcal{C}_c$ represent the memory consumption of model parameters, optimizers, activations and the CUDA context, respectively.

(1) Model parameters include all weights and biases used during fine-tuning, encompassing both trainable and frozen parameters. Based on the federated fine-tuning problem formulation in~\eqref{eq:ffm_obj}, 
the memory consumption for storing model parameters in bytes can be expressed as $\mathcal{C}_p:=\varphi(\{\boldsymbol{\theta}, \boldsymbol{\Theta}^0\}) $ where $ \varphi$ is defined as $\varphi(\boldsymbol{z}) =\eta N_{weight}(\boldsymbol{z})+\eta N_{bias}(\boldsymbol{z})$. Here, $N_{weight}(\cdot)$ and $N_{bias}(\cdot)$ are the number of weights and bias in the model $\boldsymbol{z}$, and $\eta$ is the number of bytes per parameter, which is determined by the precision (e.g., 2 bytes for 16-bit precision). 

(2) Modern gradient-based optimizers like Adam~\cite{kingma2014adam} are widely used in FM fine-tuning. They maintain state information such as first-order and second-order gradient estimates for each trainable parameter. 
Consequently, the memory consumption for optimizer states scales linearly with the number of trainable model parameters and can be approximated as $\mathcal{C}_o:=b\zeta\varphi(\boldsymbol{\theta})$, where $\zeta$ represents the number of optimizer states and $b$ is the batch size.

(3) 
Unlike optimizer states, activation memory in transformer model fine-tuning grows non-linearly with the number of trainable parameters~\cite{ardakani2024slimfit}. We distinguish between dynamic activations, which can be discarded by freezing certain parameters, and static activations, retained due to non-linear functions like GELU, LayerNorm, and Softmax (see {{Supplement~\ref{subsec:appendix-dynamic-activation}-\ref{subsec:appendix-memory-footprint-example}}} for details). We calculate the size of dynamic and static activations for each transformer block in $\{\boldsymbol{\mathcal{\theta}}, \boldsymbol{\Theta}^0\}$, when pre-trained weights $\boldsymbol{\Theta}^0$ are frozen and the LoRA modules $\boldsymbol{\mathcal{\theta}}$ are trainable. As a result, the memory consumption for storing activations during the update of $\boldsymbol{\mathcal{\theta}}$ is given by: $\mathcal{C}_a:=b\eta{\textstyle \sum_{j=1}^{l}}\sigma_{(j)}^d + b\eta{\textstyle \sum_{j=1}^{l}}\sigma_{(j)}^s$, where $\sigma_{(j)}^d$ and $\sigma_{(j)}^s$ represents the size of dynamic and static activations for the $j$-th block.

(4) Finally, the memory consumption of the CUDA context includes GPU kernel launches, memory management, and other operational resources. 
This component highly depends on the environment, machine, and configuration, thus, we treat $\mathcal{C}_c$ as a constant. Overall, the total memory consumption $\mathcal{C}$ is formulated as, 
\begin{equation}\label{eq:memory_model}
\begin{aligned}
\mathcal{C} = & \ \varphi(\{\boldsymbol{\theta}, \boldsymbol{\Theta}^0\}) + b\zeta\varphi(\boldsymbol{\theta}) + b\eta{\textstyle \sum_{j=1}^{l}}\sigma_{(j)}^d  \\
& + b\eta{\textstyle \sum_{j=1}^{l}}\sigma_{(j)}^s
 + \mathcal{C}_c.
\end{aligned}
\end{equation}
This indicates that reducing the size of $\boldsymbol{\theta}$ can further lower the memory cost, similar to existing methods~\cite{cho2024heterogeneous,wangflora,bai2024federated} that reduce the ranks of LoRA modules to save memory cost. However, this saving is limited. As shown in Figure.\ref{fig:motivation}(b), the overall memory cost is primarily dominated by activations. Reducing the ranks of LoRA modules has a minimal impact on the size of both dynamic and static activations. 

\subsection{Reducing Memory Consumption with Partial LoRA Allocation}\label{subsec:lora-allocation}
%
%

To save GPU memory consumption, the primary contributor—activation memory cost—should be reduced. 
In fact, we can eliminate the need to store the dynamic activations for a transformer block by freezing its LoRA modules, as the entire block is frozen in this case. This is why FedRA~\cite{su2023fedra}, which randomly freezes LoRA modules in some transformer blocks, can reduce certain activation memory consumption. However, previous works do not clearly model the impact of freezing LoRA modules on overall memory cost reduction. 

Let $m:=\{m_{(1)}, m_{(2)}, \cdots, m_{(l)}\}$ represent the LoRA allocation map for the model $\{\boldsymbol{\mathcal{\theta}}, \boldsymbol{\Theta}^0\}$, where $m_{{(j)}}=1$ if the LoRA module in $j$-th block, i.e., $\boldsymbol{\mathcal{\theta}}_{(j)}$, is trainable, otherwise $m_{{(j)}}=0$ if it is frozen. Let $\mathcal{J} = \{ j \mid m_{(j)} = 1,\forall j\}$, the overall memory computation with partial LoRA allocation should be: 
\begin{equation}\label{eq:memory_model2}
    \mathcal{C}(m)=\mathcal{C}_p +\mathcal{C}_o (m) +\mathcal{C}_a (m)+\mathcal{C}_c, 
\end{equation} where $\mathcal{C}_{o} (m) =  b\zeta \varphi(\{\boldsymbol{\mathcal{\theta}}_{(j)}\}_{j\in\mathcal
J})$ and
\begin{equation}\label{eqn:ca}
\mathcal{C}_{a} (m) = b\eta\sum_{j \in \mathcal{J}} \sigma_{(j)}^{d}
 + {b\eta\sum_{j=\min(\mathcal{J})
}^{l}} \sigma_{(j)}^{s}. 
\end{equation}
Freezing LoRA modules affects only the memory cost of optimizer states and activations. The memory consumption of dynamic activations (the first term in \eqref{eqn:ca}) is associated only with trainable LoRA modules. The memory consumption of static activations (the second term in \eqref{eqn:ca}) depends on the earliest trainable LoRA modules, i.e., $\min(\mathcal{J})$, since static activations from all layers after $\min(\mathcal{J})$-th layer must be stored for backpropagation~\cite{ardakani2024slimfit}. This formulation indicates that, in addition to reducing memory usage for optimizer states and dynamic activations, partial LoRA allocation has the potential to reduce the memory usage for static activations, though it must be carefully designed. In the following, we discuss two naive allocation strategies and analyze their memory consumption. 

\textit{Two Naive Cases: Memory Saver and Memory Hogger.} 
Memory Saver makes the last $u$ LoRA modules trainable, resulting in the LoRA allocation map is $m_{\text{MS}}= \{0 \ \text{for} \ j \in [1, l-u], \ 1 \ \text{for} \ j \in [l-u+1, l]\}$. Memory Hogger makes the first $u$ LoRA modules trainable, yielding the LoRA allocation map $m_{\text{MH}}= \{1 \ \text{for} \ j \in [1, u], \ 0 \ \text{for} \ j \in [u+1, l]\}$. Despite both having the same number of trainable LoRA modules, these strategies differ in their memory-saving capabilities, particularly in reducing static activation memory, as $\mathcal{C}_{\text{MH}}-\mathcal{C}_{\text{MS}}={b\eta\sum_{j=1
}^{l-u}} \sigma_{(j)}^{s}$, which can be substantial as demonstrated in Figure~\ref{fig:motivation}(a).

\section{Our Fed-pilot Framework for Federated Fine-Tuning over Heterogeneous Clients}\label{framework}
\begin{figure}[t]
\centering
\includegraphics[width=0.40\textwidth]{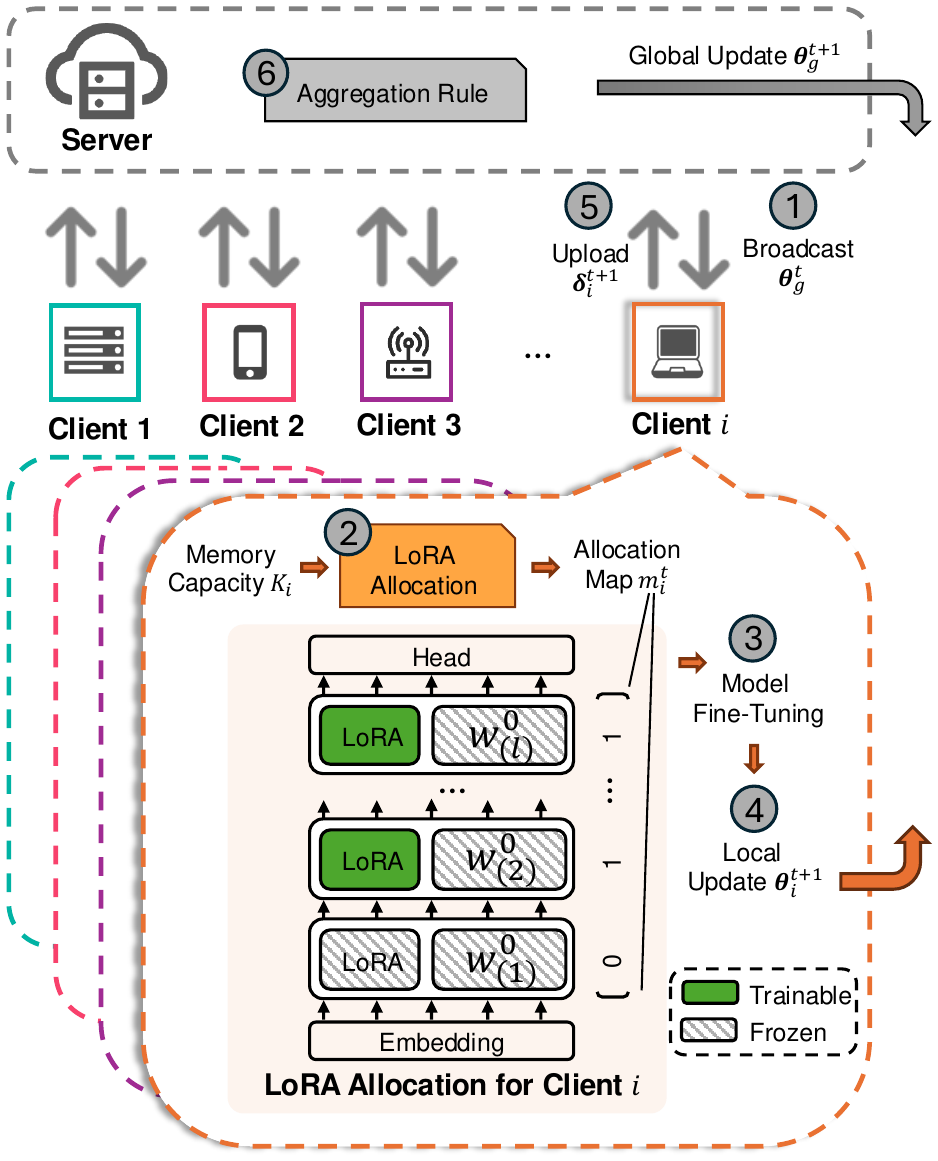}
\caption{The overall framework of Fed-pilot with optimal LoRA allocation over heterogeneous clients.}
\label{fig:framework}
\end{figure}

In this section, we introduce Fed-pilot, a novel federated fine-tuning framework that allocates partial LoRA modules for clients with limited memory capacities to perform local training. Specifically, let $\mathcal{K}_i$ represent the memory capacity of client $i$, and assume each client has its own LoRA allocation map $m_i$. The map $m_i$ is designed such that the memory consumption, i.e., $\mathcal{C}(m_i)$, does not exceed the client's memory capacity $\mathcal{K}_i$. In Fed-pilot, clients have a local copy of the pre-trained weights $\boldsymbol{\Theta}^0$ and aim to collaboratively train the global LoRA modules $\boldsymbol{\mathcal{\theta}}$ following these steps: {(1) At each round $t$, the server randomly selects a subset of clients $\mathcal{S}^t$ and broadcasts the current global LoRA module $\boldsymbol{\mathcal{\theta}}^t$ to them.} 
Each selected client initializes its model as $\{\boldsymbol{\mathcal{\theta}}_i^t, \boldsymbol{\Theta}^0 \}$ with $\boldsymbol{\mathcal{\theta}}_i^t = \boldsymbol{\mathcal{\theta}}^t$; (2) Each selected client determines its optimal LoRA allocation map $m_i^t$ based on its memory capacity $\mathcal{K}_i$ and freezes the LoRA module in the $j$-th transformer block, i.e., $\boldsymbol{\mathcal{\theta}}_{i, (j)}^t$, if $m_{i, (j)}^t=0$; (3) Selected clients update only their trainable LoRA modules to obtain $\boldsymbol{\mathcal{\theta}}_i^{t+1}$; (4) Clients obtain their model updates, $ \boldsymbol{\delta}_i^{t+1} := \boldsymbol{\mathcal{\theta}}_i^{t+1} - \boldsymbol{\mathcal{\theta}}_i^t$; (5) Clients upload the model updates to the server; 
 (6) Server aggregates all the received updates and computes a new global model $\boldsymbol{\mathcal{\theta}}^{t+1}$ for the next round of training. This process iterates for $T$ rounds until the fine-tuned model $\{\boldsymbol{\mathcal{\theta}}^T, \boldsymbol{\Theta}^0\}$ converges. We illustrate the pipeline of our Fed-pilot framework in Figure~\ref{fig:framework}.

To ensure fine-tuning performance under the memory constraints, we optimize the LoRA allocation using knapsack optimization (Section~\ref{sec:optlora}) and design a new aggregation rule for the heterogeneous local model updates (Section~\ref{subsec:HAgg}). 

\subsection{Optimizing LoRA Allocation under Heterogeneous Memory Constraints}\label{sec:optlora}

\subsubsection{Knapsack Optimization Problem}


The knapsack optimization problem aims to maximize the total \textit{value} of selected \textit{items} while ensuring that their combined \textit{weight} does not exceed a predefined \textit{capacity}. In this context, \textit{value} corresponds to the utility or benefit of an {item}, while \textit{weight} represents its associated cost or resource consumption. The \textit{capacity} defines the maximum allowable total {weight} for the selected items.

In our setting, we employ knapsack optimization to maximize fine-tuning performance by optimizing the LoRA allocation map for each client under memory constraints. Specifically, for every client $i\in[v]$, we define the following constrained optimization problem: 
\begin{equation}\label{eq:optimization}
\max_{m_{i}} \sum_{j\in[l]}m_{i,(j)}{G}(\boldsymbol{\mathcal{\theta}}_{i, (j)}), \text{ subject to } \mathcal{C}(m_i) \leq\mathcal{K}_i.
\end{equation}
Here, ${G}(\cdot)$ is the value function, defined in \eqref{eq:valuefunction}, which quantifies the contribution of the LoRA module within a specific transformer block to the global model's performance.  $\mathcal{C}(\cdot)$ is the overall weight function, defined in \eqref{eq:memory_model2}, which measures the total memory consumption for a given allocation map. $\mathcal{K}_i$ denotes the memory capacity of client $i$. 

The problem is NP-hard in the discrete setting. The classical heuristic approach to solving it is the greedy algorithm, which iteratively selects items based on their value-to-weight ratio until reaching capacity. However, unlike standard knapsack optimization, where item weights are static, the weight of each LoRA module $\boldsymbol{\mathcal{\theta}}_{i, (j)}$ dynamically adapts based on the previously assigned LoRA modules. Specifically, at each step $h\in[0, l-1]$, the weight of an item $\boldsymbol{\mathcal{\theta}}_{i, (j)}$ is $\mathcal{C}_{i,(j)}^{h} = \mathcal{C}(m_i^h)$ if $h=0$, otherwise, 
\begin{multline}\label{eq:weightfunction}
\mathcal{C}_{i,(j)}^{h} = b\zeta\varphi(\boldsymbol{\mathcal{\theta}}_{i, (j)}) + b\eta\sigma_{(j)}^{d} \notag  \\
 + b\eta\sum_{j^{\prime}=\min(\mathcal{J}^{h-1} \cup \{ j \})}^{l} \sigma_{(j^\prime)}^{s} - b\eta\sum_{j^{\prime}=\min(\mathcal{J}^{h-1})}^{l} \sigma_{(j^\prime)}^{s},
\end{multline}
where $m_{i}^{h}$ represents the allocation map at step $h$ and $\mathcal{J}^{h-1} = \{ j \mid m_{i,(j)}^{h-1} = 1, \forall j \}$. Note that these item weights will be normalized using min-max scaling at every step.

\subsubsection{Information Gain Score based Value Function}\label{sec:IGScore}

Here, we define the value function used in our LoRA allocation optimization problem in \eqref{eq:optimization}, which quantifies the contribution of each LoRA module to fine-tuning performance.
From an information-theoretic perspective, mutual information quantifies how much knowledge of one variable (e.g., a model parameter) reduces the uncertainty of another (e.g., task-specific loss) \cite{kullback1997information}. This suggests that a layer's importance in fine-tuning can be measured by how much information it contributes to minimizing the loss \cite{gabrie2018entropy}. However, directly computing mutual information in deep learning models is often intractable due to its high computational complexity.

To make this concept computationally feasible, we approximate a layer’s contribution using its gradient norm, which reflects how much a parameter influences loss reduction. A higher gradient norm indicates that a layer plays a more active role in optimization, capturing critical information necessary for improving model performance. Based on this intuition, we introduce the IG score $I_{(j)}$, a gradient-based metric that serves as a practical proxy for evaluating the information gain of the $j$-th LoRA module. 
\begin{equation}\label{eq:igscore}
I_{(j)}(\boldsymbol{\mathcal{\theta}}, D_{\text{IG}})=\sum^{ \lceil|D_{\text{IG}}|/|B^e_{\text{IG}}| \rceil}_{e=1}\left\|\nabla \ell(\boldsymbol{\mathcal{\theta}}; \boldsymbol{\Theta}^0, B_{\text{IG}}^e)_{(j)}\right\|_2^2,
\end{equation}
where $D_{\text{IG}}$ represents an IG dataset used to estimate the impact of each LoRA module, and $B_{\text{IG}}^{e}$ is a mini-batch sampled from $D_{\text{IG}}$ at the $e$-th forward-backward iteration.

However, calculating this IG score is challenging in federated settings. The server must obtain an IG dataset that aligns with the distribution of local datasets~\cite{zhu2021data}. However, if clients compute the IG score locally, the results may be sparse, when some LoRA modules are frozen and do not contribute to the gradient computation. Furthermore, when clients have Non-IID data distributions, a globally computed IG score may fail to reflect the client-specific importance of LoRA modules, leading to suboptimal LoRA allocation strategies. 

To address these challenges of sparsity and data heterogeneity, we design a three-step IG score aggregation process. Specifically, at round $t$, each client computes the IG score for its local trainable LoRA modules 
using a small subset $D_i^{\text{IG}}$ extracted from its local dataset, where the IG score for the $j$-th LoRA module is calculated by $I_{i,(j)}^t = I_{(j)} (\boldsymbol{\mathcal{\theta}}_i^{t}, D_i^{\text{IG}}) $ if $m_{i, (j)}^t = 1$.  
Let $\mathcal{V}_{(j)}^t = \{i; m_{i, (j)}^t = 1, i\in\mathcal{S}^t\}$ to represent the set of clients that train the $j$-th LoRA module at round \( t \) and $\mathcal{T}_{(j)}^t = \{t^\prime; \mathcal{V}_{(j)}^{t^\prime}\neq \emptyset, t^\prime \in (t-T_{\text{IG}}, t]\}$ to represent the set of past rounds in which at least one client trains the $j$-th LoRA module within the most recent \( T_{\text{IG}} \) rounds. Note that the temporal aggregation window$T_{\text{IG}}$ is selected to ensure that $\mathcal{T}_{(j)}^{t}\neq \emptyset$. Based on these, we define the value function of client $i$ as follows
\begin{align}\label{eq:valuefunction}
    {G}(\boldsymbol{\mathcal{\theta}}_{i, (j)}^t) = \frac{I_{i,(j)}^t + \frac{1}{|\mathcal{T}_{(j)}^t|}\sum_{t^\prime\in \mathcal{T}_{(j)}^t} \bar{I}_{(j)}^{t^\prime}}{{m_{i, (j)}^t + 1}} \nonumber\\ \text{ with } \bar{I}_{(j)}^{t^\prime} = \frac{1}{|\mathcal{V}_{(j)}^{t^\prime}|}\sum_{i\in \mathcal{V}_{(j)}^{t^\prime}} I_{i,(j)}^t.
\end{align}

To mitigate fluctuations and address sparsity in local scores, the value function first performs {cross-client aggregation} to compute an averaged IG score for each LoRA module. Then, it applies {temporal averaging} over the most recent \( T_{\text{IG}} \) rounds to further stabilize the score. Finally, to account for the unique data distribution of each client, the {temporally smoothed average score} is combined with the {client’s local score}, ensuring a balance between global knowledge and local adaptation in determining the final value of a LoRA module. 

\subsection{Aggregation with Update Compensation}\label{subsec:HAgg}

With optimized LoRA allocation, clients with varying memory capacities train different sets of LoRA modules, leading to diverse model updates. From a global perspective, this divergence causes variations in both the frequency and the {scale} of the parameter changes across LoRA modules. Specifically, some modules receive updates from a larger number of clients, while others receive updates from fewer clients. As a result, due to the Non-IID nature of client data, different layers may learn from distinct data distribution. In this case, simply averaging model updates can hinder convergence. 

To address this issue, we design a compensated aggregation rule ({ComAgg}), which dynamically compensates the layers that are less updated at the current round. Specifically, upon received the local model updates $\{\boldsymbol{\delta}_i^{t+1}\}_{i\in \mathcal{S}^t}$ from the selected clients at round $t$, the server aggregates them layer-wise as follows: for each $j\in[l]$,
\begin{align}\label{eq:agg_temporal}
    {\boldsymbol{\delta}}_{(j)}^{t+1} =  \frac{\beta_{(j)}^t}{\alpha_{(j)}^t+\beta_{(j)}^t}\boldsymbol{\delta}_{(j)}^{t} + \frac{\alpha_{(j)}^t}{\alpha_{(j)}^t+\beta_{(j)}^t} \bar{\boldsymbol{\delta}}_{(j)}^{t+1} \nonumber\\
    \text{ with } \bar{\boldsymbol{\delta}}_{(j)}^{t+1} = \frac{1}{|\mathcal{S}_{(j)}^t|} \sum_{i \in \mathcal{S}_{(j)}^t} {\boldsymbol{\delta}}_{i, (j)}^{t+1},
\end{align}
where $\mathcal{S}_{(j)}^t = \{i; m_{i, (j)}^t=1, i\in \mathcal{S}^t\}$ represents the set of clients that train the $j$-th LoRA module at round $t$. The coefficient $\alpha_{(j)}^t=|\mathcal{S}_{(j)}^t|$ represents the number of clients training the $j$-th LoRA module at round $t$, and $\beta_{(j)}^t=\frac{1}{ T_{\text{Agg}}} \sum_{t^\prime = t - T_{\text{Agg}} + 1}^{t} \alpha_{(j)}^{t^\prime}$ captures the average number of clients contributing to the $j$-th LoRA module over the past $T_{\text{Agg}}$ rounds. By compensating under-updated LoRA modules using historical global updates, ComAgg mitigates inconsistencies in data distribution across modules, ensuring more stable convergence.

\begin{table}[t]
  \centering
  \caption{Main results on CIFAR-100 dataset with ViT-base model in both IID and Non-IID settings.}
  \scalebox{0.88}{
    \begin{tabular}{c|cccc}
    \toprule[1pt]
    \multirow{2}{*}{\textbf{Methods}} & \multicolumn{4}{c}{\textbf{Model Performance (Accuracy)}} 
    \\
    \cline{2-5}
    & \textbf{IID} & \textbf{Non-IID (20)} & \textbf{Non-IID (10)} & \textbf{Avg} \\
    \midrule
    Baseline-EL & 40.15 & 27.82 & 7.04  & 25.00 \\
    Baseline-MH & 53.36 & 29.77 & 7.92  & 30.35 \\
    Baseline-MS & 76.38 & 68.34 & 57.63 & 67.45 \\ \midrule
    FLoRA &1.61 &0.90 &0.78 &1.09 \\
    FlexLoRA &1.84 &1.92 &1.86 &1.87 \\
    HETLoRA &6.91 &8.23 &1.59 &5.57 \\
    FedRA & 72.60 & 62.95 & 47.98 & 61.17 \\
    \midrule
    \textbf{Ours} & \cellcolor[rgb]{ .91,  .91,  .91}80.09 & \cellcolor[rgb]{ .91,  .91,  .91}73.35 & \cellcolor[rgb]{ .91,  .91,  .91}63.93 & \cellcolor[rgb]{ .91,  .91,  .91}72.45 \\
    \bottomrule[1pt]
    \end{tabular}%
    }
  \label{tab:cifar100-results}%
\end{table}

\begin{table}[t]
  \centering
  \caption{Main results on LEDGAR dataset with BERT-base model in both IID and Non-IID settings.}
  \scalebox{0.88}{
    \begin{tabular}{c|cccc}
    \toprule[1pt]
    \multirow{2}{*}{\textbf{Methods}} & \multicolumn{4}{c}{\textbf{Model Performance (MacroF1)}} 
    \\
    \cline{2-5}
    & \textbf{IID} & \textbf{Non-IID (20)} & \textbf{Non-IID (10)} & \textbf{Avg} \\
    \midrule
    Baseline-EL & 60.76 & 43.94 & 0.12  & 34.94 \\
    Baseline-MH & 61.58 & 6.59  & 0.22  & 22.79 \\
    Baseline-MS & 62.95 & 56.22 & 46.93 & 55.36 \\ \midrule
    FLoRA &0.10 &0.07 &0.01 &0.06 \\
    FlexLoRA &0.10 &0.24 &0.32 &0.22 \\
    HETLoRA &0.51 &0.32 &0.10 &0.31 \\ 
    FedRA & \cellcolor[rgb]{ .91,  .91,  .91}63.20 & 54.96 & 47.24 & 55.13 \\
    \midrule
    \textbf{Ours} & 63.17 & \cellcolor[rgb]{ .91,  .91,  .91}56.31 & \cellcolor[rgb]{ .91,  .91,  .91}48.87 & \cellcolor[rgb]{ .91,  .91,  .91}56.11 \\
    \bottomrule[1pt]
    \end{tabular}%
    }
  \label{tab:ledgar-results}%
\end{table}


\section{Evaluation}\label{evaluation}
\subsection{Experimental Settings}
\subsubsection{Memory Capacity Heterogeneity}
%
%
We define four memory capacity levels (Level 1-4) based on real-world NVIDIA GPUs. For FMs with millions of parameters, we set the client memory capacities to 24 GB, 32 GB, 40 GB, and 48 GB for Level 1 through 4, respectively. For FMs with billions of parameters, the corresponding memory capacities are to 32 GB, 40 GB, 48 GB, and 80 GB. See {{Supplement Section~\ref{subsec:appendix-capacity-settings}}} for detailed capacity settings on different models. We assume that clients are distributed across these levels in a 4:3:2:1 ratio, from the lowest to the highest memory capacity, reflecting the higher prevalence of low-memory clients in real-world scenarios.


\subsubsection{Data Heterogeneity}

We evaluate our approach on five distinct datasets while simulating varying levels of Non-IID settings, including category heterogeneity, task heterogeneity, and domain heterogeneity.

\textbf{Category Heterogeneity.} We use CIFAR-100 dataset~\cite{krizhevsky2009learning} and LEDGAR dataset from LexGLUE~\cite{chalkidis-etal-2021-lexglue} to simulate the category heterogeneity settings by skewing the label distribution across clients. For the pre-trained model, we use ViT-base~\cite{dosovitskiy2020image} for visual tasks and BERT-base~\cite{devlin2019bert} for language tasks. 

\textbf{Task Heterogeneity.} We use Natural Instruction (NI) dataset~\cite{wang2022super} and Dolly-15K~\cite{conover2023free} dataset that contain diverse natural language tasks to evaluate our approach under task heterogeneity. 
We use a Llama-like model Data-Juicer-1B~\cite{chen2024data} and the OPT-1.3B~\cite{zhang2022opt} as the pre-trained models for these two datasets, respectively. 

\textbf{Domain Heterogeneity.} 
We use a customized DomainNet dataset~\cite{peng2019moment} (referred to as DomainNet-121), which contains six visual domains, to simulate the domain heterogeneity by assigning data from only one domain to each client. For this dataset, we use ViT-base~\cite{dosovitskiy2020image} as the pre-trained model. 

We provide the details of data settings in {{Supplement Section~\ref{sec:appendix-dataset}}}.

\subsubsection{Baselines}
To evaluate the effectiveness of Fed-pilot, we implement seven baselines:

\textit{Baseline-Exclusive Learning (EL)} excludes clients that cannot train all LoRA modules. 
\textit{Baseline-Memory Saver (MS)} allocates the last $u$ LoRA modules according to the capacity, prioritizing the training of more modules.
%
\textit{Baseline-Memory Hogger (MH)} allocates the first $u$ LoRA modules according to the capacity,
emphasizing the utilization of shallow-layer features.
%
{HETLoRA}~\cite{cho2024heterogeneous} is a rank-based method designed to accommodate heterogeneous clients with varying LoRA ranks.
{FLoRA}~\cite{wangflora} assigns lower ranks to resource-constraint clients and shares stacked LoRA modules for updates.
{FlexLoRA}~\cite{bai2024federated} improves aggregation in rank-based methods using SVD.
{FedRA}~\cite{su2023fedra} 
randomly allocates a subset of LoRA modules for clients with limited memory capacities to train at each round. 

\subsubsection{Implementation Details}
We use PyTorch to implement our framework. All experiments are carried out on a server with 8 $\times$ NVIDIA RTX A6000 GPUs. The total GPU hours for running all the experiments are over 3,000 GPU hours. For the configuration of models,  hyper-parameters, and FL settings, 
please refer to {{Supplement Section~\ref{sec:appendix-settings}}} for more details.


\begin{figure}[t]
\centering
\includegraphics[width=0.48\textwidth]{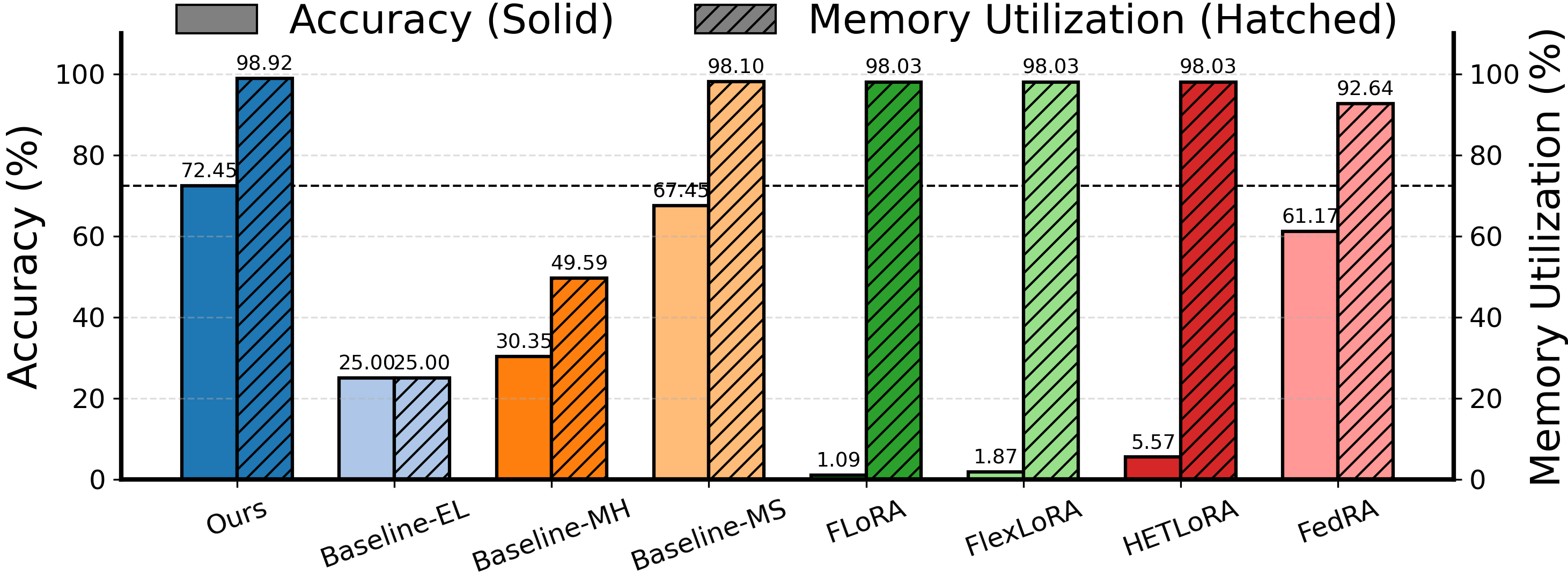}
\caption{The average accuracy vs. average memory utilization across methods on CIFAR-100 dataset.}
\label{fig:memory_utilization}
\end{figure}

\begin{figure*}[t]
\centering
\includegraphics[width=0.95 \textwidth]{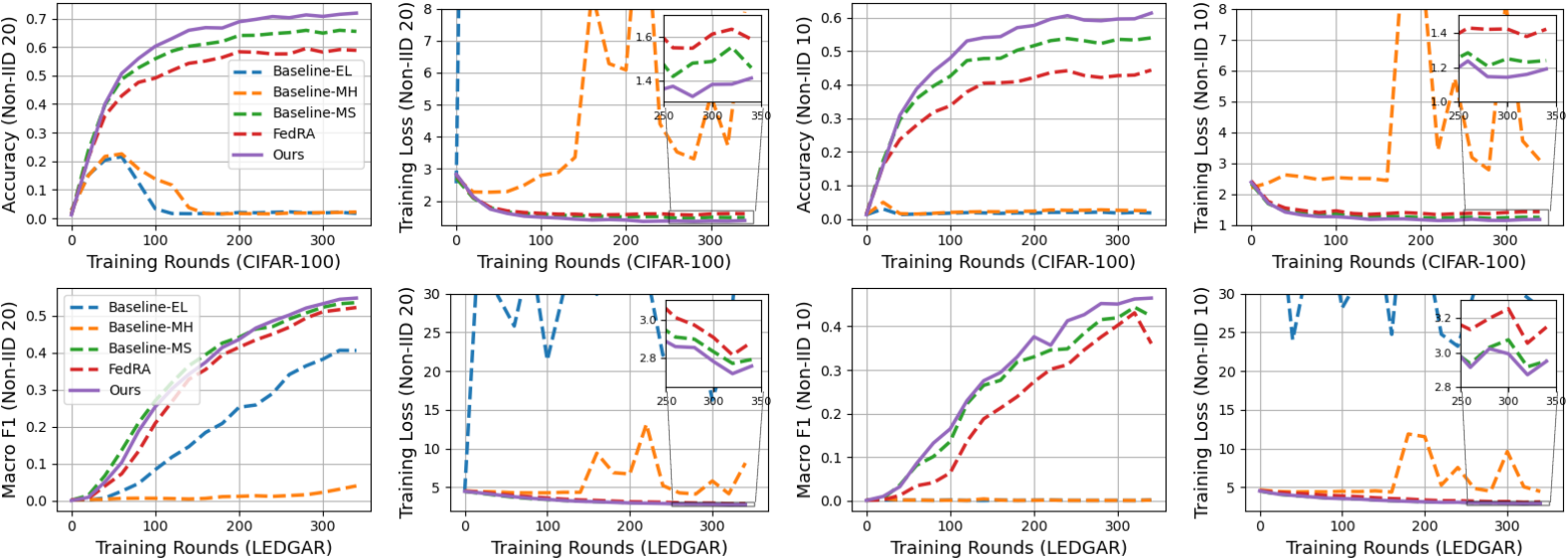}
\caption{Convergence performance on CIFAR-100 and LEDGAR datasets. Rank-based methods are excluded from the results as they diverge.}
\label{fig:cifar100-ledgar-metric-loss}
\end{figure*}

\subsection{Experimental Results}

The best results are highlighted in \colorbox[rgb]{ .91,  .91,  .91}{grey}. 

\begin{table}[t]
  \centering
  \caption{Main results on NI dataset with Llama-1B (Data-Juicer) model in both IID and Non-IID settings.}
  \scalebox{0.88}{
    \begin{tabular}{c|cccc}
    \toprule[1pt]
    \multirow{2}{*}{\textbf{Methods}} & \multicolumn{4}{c}{\textbf{Model Performance (Rouge-L)}} 
    \\
    \cline{2-5}
    & \textbf{IID} & \textbf{Non-IID (20)} & \textbf{Non-IID (15)} & \textbf{Avg} \\
    \midrule
    Baseline-EL & 57.95 & 56.28 & 42.77 & 52.33 \\
    Baseline-MH & 58.58 & 58.08 & 56.78 & 57.81 \\
    Baseline-MS & 58.14 & 57.80 & 58.03 & 57.99 \\\midrule
    FLoRA & 42.63 & 40.84 & 37.72 & 40.39        \\
    FlexLoRA & 36.72 & 34.15 & 34.45 & 35.10    \\
    HETLoRA & 54.92 & 58.28 & 57.83 & 57.01     \\
    FedRA & 58.18 & 57.10 & 58.13 & 57.80       \\
    \midrule
    \textbf{Ours} & \cellcolor[rgb]{ .91,  .91,  .91}58.73      & \cellcolor[rgb]{ .91,  .91,  .91} 58.59     & \cellcolor[rgb]{ .91,  .91,  .91} 59.30     & \cellcolor[rgb]{ .91,  .91,  .91} 58.87     \\
    \bottomrule[1pt]
    \end{tabular}%
    }
  \label{tab:natural-instruction}%
\end{table}%

\textbf{Results on CIFAR-100.} Table~\ref{tab:cifar100-results} presents the main results on CIFAR-100 dataset. We report the Top-1 accuracy of the fine-tuned global model under both IID and two different Non-IID (20, 10) settings. Our method demonstrates substantial improvements over the baseline methods. 
Compared to Baseline-MS, our method achieves improvements of +3.71\%, +5.01\%, and +6.30\% across the three data distributions, respectively. This highlights the effectiveness of our proposed LoRA allocation optimization, indicating that fine-tuning as many LoRA modules as possible is not always beneficial.
Our method also outperforms FedRA by an average of +11.28\%, as FedRA relies on random sampling without considering either memory utilization or the contribution of each LoRA module to model accuracy.
Compared to Baseline-EL and Baseline-MH, our method achieves average performance gains of +47.45\% and +42.10\%, respectively, and demonstrates much more stable performance under Non-IID settings. This is because these two methods cannot accommodate clients with low memory capacity.
Rank-based methods such as FLoRA, FlexLoRA, and HETLoRA diverge due to inefficient memory management, and extreme low LoRA ranks.

\textbf{Results on LEDGAR.} Table~\ref{tab:ledgar-results} presents the main results on LEDGAR dataset. We report Macro F1 score of the fine-tuned global model under both IID and Non-IID (20, 10) settings.
Compared to FedRA, our method achieves an average improvement of +0.98\%. Although it shows a slight drop under the IID distribution for our method compared to FedRA, it excels in handling Non-IID settings, with gains of +1.35\% and +1.63\% under ``Non-IID (20)'' and ``Non-IID (10)'' settings, respectively. These results highlight the adaptability of our framework to heterogeneous data distributions. 
Our method also outperforms Baseline-MS by +1.94\% under the ``Non-IID (10)'' setting.
Baseline-EL and Baseline-MH shows a significant decline compared to our method, underscoring the importance of incorporating comprehensive client data information. Rank-based methods such as FLoRA, FlexLoRA, and HETLoRA still diverge when clients have extremely low LoRA ranks.

\textbf{Results on NI.} Table~\ref{tab:natural-instruction} presents the main results on the NI dataset. We report the Rouge-L score of the fine-tuned global model under both IID and two different Non-IID (20, 15) settings. 
%
Our method demonstrates significant improvements over rank-based baselines, achieving average Rouge-L score gains of +18.48\%, +23.77\%, and +1.86\% compared to FLoRA, FlexLoRA, and HETLoRA, respectively. These improvements are attributed to the inefficiency of rank-based methods in memory management, which restricts the rank of LoRA modules. Additionally, redistributed methods such as stacking and SVD can become unstable when the rank of LoRA modules is extremely low.
Compared to Baseline-EL and Baseline-MH, our method achieves improvements of +16.53\% and +2.52\%, respectively, under the Non-IID (15) setting, indicating strong adaptability to heterogeneous data distributions.
Our method also consistently maintains stable performance gains across all data distributions when compared to Baseline-MS and FedRA. Notably, under the extreme Non-IID setting with 15 tasks per client, it outperforms these baselines by +1.27\% and +1.17\%, respectively.
We would like to point out that the task-level Non-IID skew in natural language datasets does not fully reflect the effectiveness of task heterogeneity, which affects trends of the Rouge-L scores reported in the table.
%


\begin{table}[t]
  \centering
  \caption{Main results on Dolly-15K dataset with OPT-1.3B model in both IID and Non-IID settings.}
  \scalebox{0.88}{
    \begin{tabular}{c|cccc}
    \toprule[1pt]
    \multirow{2}{*}{\textbf{Methods}} & \multicolumn{4}{c}{\textbf{Model Performance (Rouge-L)}} 
    \\
    \cline{2-5}
    & \textbf{IID} & \textbf{Non-IID (3)} & \textbf{Non-IID (1)} & \textbf{Avg}  \\
    \midrule
    Baseline-EL & 57.99 & 59.05 & 57.52 & 58.18  \\
    Baseline-MH & 58.39 & 58.33 & 56.07 & 57.59 \\
    Baseline-MS & 58.14 & 58.01 & 58.00 & 58.05 \\\midrule
    FLoRA & 40.33 & 40.49 & 40.37 & 40.39       \\
    FlexLoRA & 40.12 & 40.13 & 40.21 & 40.15    \\
    HETLoRA & 58.76 & 59.17 & 57.80 & 58.57     \\
    FedRA & 58.21 & 58.64 & 57.94 & 58.26      \\
    \midrule
    \textbf{Ours} & \cellcolor[rgb]{ .91,  .91,  .91}59.46     & \cellcolor[rgb]{ .91,  .91,  .91}59.36     & \cellcolor[rgb]{ .91,  .91,  .91}59.37     & \cellcolor[rgb]{ .91,  .91,  .91}59.39     \\
    \bottomrule[1pt]
    \end{tabular}%
    }
  \label{tab:dolly-15k}%
\end{table}%

\begin{table}[t]
  \centering
  \caption{Main results on DomainNet-121 dataset with ViT-base model in both IID and Non-IID settings.}
  \scalebox{0.88}{
    \begin{tabular}{c|cccc}
    \toprule[1pt]
    \multirow{2}{*}{\textbf{Methods}} & \multicolumn{4}{c}{\textbf{Model Performance (Accuracy)}} 
    \\
    \cline{2-5}
    & \textbf{IID} & \textbf{Non-IID (50)} & \textbf{Non-IID (25)} & \textbf{Avg} \\
    \midrule
    Baseline-EL & 50.33 & 38.34 & 21.15 & 36.60 \\
    Baseline-MH & 50.19 & 38.20 & 19.61 & 36.00 \\
    Baseline-MS & 58.92 & 53.60 & 40.82 & 51.11 \\ \midrule
    FLoRA &1.10 &1.04 &1.01 &1.05 \\
    FlexLoRA &0.85 &0.85 &0.88 &0.86 \\
    HETLoRA &6.94 &3.30 &0.96 &3.73 \\ 
    FedRA & 57.38 & 53.05 & 38.07 & 49.50 \\
    \midrule
    \textbf{Ours} & \cellcolor[rgb]{ .91,  .91,  .91}59.36 & \cellcolor[rgb]{ .91,  .91,  .91}54.40 & \cellcolor[rgb]{ .91,  .91,  .91}42.20 & \cellcolor[rgb]{ .91,  .91,  .91}51.98 \\
    \bottomrule[1pt]
    \end{tabular}%
    }
  \label{tab:domainnet-results}%
\end{table}

\textbf{Results on Dolly-15K.}
Table~\ref{tab:dolly-15k} presents the main results on Dolly-15K dataset. We report Rouge-L scores of the fine-tuned global model under both IID and Non-IID (3, 1) settings. From the results, our method achieves the highest average Rouge-L score of 59.39\%, outperforming all baselines. Notably, our approach maintains stable Rouge-L scores above 59\% across all data distributions.
In contrast, rank-based methods FLoRA and FlexLoRA achieve around 40\% accuracy, likely due to their noise-alleviation aggregation being ineffective when some LoRA modules with extreme low rank. Among them, FlexLoRA exhibits the lowest average Rouge-L score of 40.15\%.
%
%
Among the baselines, Baseline-EL, Baseline-MH, Baseline-MS, and FedRA perform similarly, with average scores ranging from 57.59\% to 58.26\%.

\textbf{Results on DomainNet-121.} 
Table~\ref{tab:domainnet-results} presents the main results on DomainNet-121. For the domain heterogeneity, each client only contain dataset from one domain. Our method shows significant improvements over the baseline methods. Specifically, our approach achieves average accuracy gains of +15.98\% over Baseline-MH, +15.38\% over Baseline-EL, and +2.48\% over FedRA. Even though Baseline-MS trains more LoRA modules, our approach still surpasses it by +0.87\% on average, demonstrating the effectiveness of our LoRA allocation strategy. This improvement remains consistent across all levels of data heterogeneity, further emphasizing the superiority of our approach. Rank-based methods diverge due to the same reasons we discussed above.

\subsection{Memory Utilization and Convergence Performance}

\textbf{Accuracy and Memory Utilization Comparisons.} {To better understand the memory utilization across methods, we visualize the average accuracy versus average memory utilization on CIFAR-100 dataset.} The average memory utilization is calculated by $\frac{1}{v}\sum_{i=1}^{v}\frac{\mathcal{C}(m_i)}{\mathcal{K}_i}$. As shown in Figure~\ref{fig:memory_utilization}, our method achieves the highest average model accuracy and memory utilization among all methods. Specifically, Baseline-EL excludes clients that cannot participate in full LoRA modules fine-tuning, resulting in only 25.00\% memory utilization. Baseline-MH focuses more on LoRA modules in shallow layers and excludes clients that cannot even fine-tune the first LoRA modules, resulting in 49.59\% memory utilization. FedRA randomly samples LoRA modules, leading to lower memory utilization compared to our memory optimization-based method. Rank-based methods such as FLoRA, FlexLoRA, and HETLoRA maintain high memory utilization but suffer from bad convergence when the LoRA ranks are too low. Baseline-MS has comparable memory utilization to ours, however, it heuristically selects as many as LoRA modules as possible in higher layers without considering their contribution. Our method outperforms Baseline-MS by +5.00\% in accuracy.

\textbf{Convergence Performance.} Figure~\ref{fig:cifar100-ledgar-metric-loss} {presents the convergence performance of baselines on CIFAR-100 and LEDGAR under two different Non-IID (20, 10) settings.} Across all settings, our method consistently outperforms baselines and converges in a stable manner. 
FedRA and Baseline-MS select random or fixed LoRA modules within memory constraints, resulting in significantly lower accuracy and Macro F1 scores compared to our approach on CIFAR-100 and LEDGAR.
Both Baseline-EL and Baseline-MH exhibit poor convergence, and Baseline-MH diverges significantly after few rounds. This indicates that limited client participation or training only a few shallow layers can lead to model divergence.

\subsection{Ablation Studies}

\begin{figure}[t]
\centering
\includegraphics[width=0.45\textwidth]{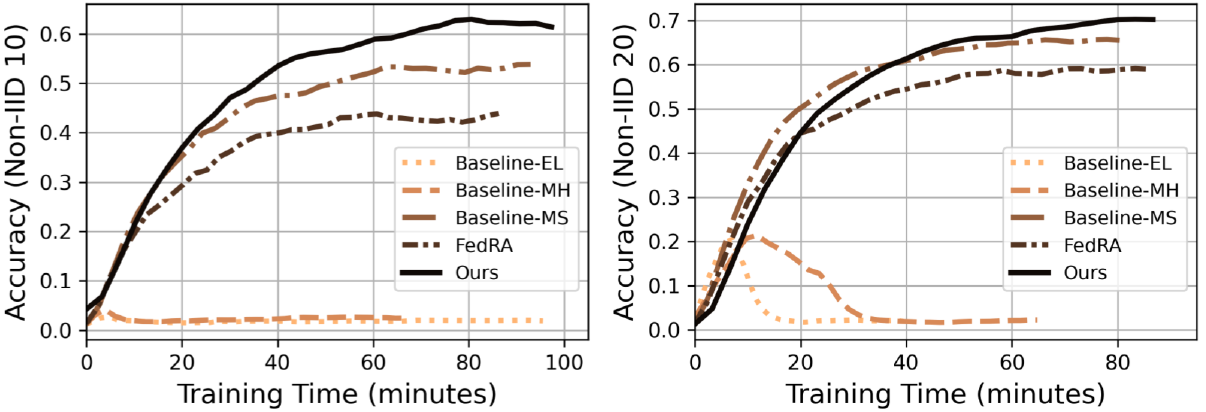}
\caption{Accuracy over training time on CIFAR-100 dataset (other methods not shown in the figure diverges).}
\label{fig:comp_eff}
\end{figure}

\begin{table}[t]
  \centering
    \caption{Ablation studies on IG score-based value function and aggregation rule ComAgg.}
    \scalebox{0.88}{
    \begin{tabular}{c|ccc}
    \toprule[1pt]
    \textbf{Methods}  & \textbf{IID} & \textbf{Non-IID (10)} & \textbf{Avg} \\
    \midrule
    Local IG score-based              & 75.81 & 58.80 & 67.30  \\
    Global IG score-based (using proxy data) & 79.74 & 63.17 & 71.45 \\ \midrule
    FedAvg & 75.86 & 57.35 & 66.60 \\
    ComAgg with $\alpha_{(j)}^t=\beta_{(j)}^t=1$ & 79.31 & 62.18 & 70.74 \\ \midrule
    \textbf{Ours}        & \cellcolor[rgb]{ .91,  .91,  .91}80.09 & \cellcolor[rgb]{ .91,  .91,  .91}63.93 & \cellcolor[rgb]{ .91,  .91,  .91}72.01 \\
    \bottomrule[1pt]
    \end{tabular}%
    }
  \label{tab:ab1}%
\end{table}%

\begin{table}[t]
  \centering
    \caption{The impact the number of total clients. Experiments are conducted on CIFAR-100 with Non-IID (10) settings. Accuracies are reported.}
    \scalebox{0.88}{
    \begin{tabular}{c|ccc}
    \toprule[1pt]
    \textbf{Method} &{\textbf{50}} &{\textbf{100}} &{\textbf{200}}  
    \\
    \midrule
    Baseline-MS & 40.52 & 57.63 & 54.90 \\
    FedRA & 26.92 &47.98 & 44.90 \\
    Ours &\cellcolor[rgb]{ .91,  .91,  .91}59.46 & \cellcolor[rgb]{ .91,  .91,  .91}63.93 & \cellcolor[rgb]{ .91,  .91,  .91}63.50  \\
    \bottomrule[1pt]
    \end{tabular}%
    }
  \label{tab:sens-num-clients}%
\end{table}%

To evaluate the effectiveness of our IG score-based value function, we conduct two ablation studies on CIFAR-100 dataset. We compare our value function with other alternations that only use the local IG score $I_i^t$ or a global IG score that is calculated using a proxy dataset sampled from the training data of clients. 

Table~\ref{tab:ab1} presents results under both IID and Non-IID (10) settings. Our method consistently outperforms others, improving over the Local IG Score-based strategy by +4.28\% and +5.13\% for IID and Non-IID, respectively. This highlights the necessity of global information. Our method also surpasses the Global IG Score-based method by +0.35\% (IID) and +0.76\% (Non-IID), demonstrating that proxy data alone is insufficient without local information.


\textbf{Aggregation Rules Evaluation.} 
We compare our aggregation rules with FedAvg~\cite{mcmahan2017communication} and ComAgg variant with coefficients $\alpha_{(j)}^t=\beta_{(j)}^t=1$. As shown in Table~\ref{tab:ab1}, our method outperforms FedAvg by +4.23\% in IID and +6.58\% in Non-IID settings. The dynamic compensation strategy further improves ComAgg by +1.27\% on average, demonstrating the benefits in mitigating model divergence.

\begin{table}[t]
\centering
\caption{The impacts of $T_{\text{IG}}$ and $T_{\text{Agg}}$. Experiments were conducted on CIFAR-100 under the Non-IID (10) setting. Each row varies one hyperparameter while fixing the other.}
\scalebox{0.88}{
\begin{tabular}{ccccc}
\toprule[1pt]
\textbf{Varying} & \textbf{Fixed Setting} & \textbf{5} & \textbf{10} & \textbf{20} \\
\hline
$T_{\text{IG}}$ & $T_{\text{Agg}}=10,\ |D_{\text{IG}}|=50$ & 63.15 & 63.93 & 62.93 \\
$T_{\text{Agg}}$ & $T_{\text{IG}}=10,\ |D_{\text{IG}}|=50$ & 63.24 & 63.93 & 63.41 \\
\bottomrule[1pt]
\end{tabular}
}
\label{tab:sens_tig_tagg}
\end{table}

\subsection{Impact of Hyperparameters}

We conduct experiments on CIFAR-100 dataset with the Non-IID (10) setting to evaluate the impact of hyperparameters such as the number of total clients, aggregation hyperparameters, and the size of the local IG dataset.

\textbf{The Number of Total Clients.} We keep the client sampling rate to 10\%, and vary the number of total clients to 50, 100, and 200. We only compare with Baseline-MS and FedRA, as other methods diverge in this Non-IID scenario. As shown in Table~\ref{tab:sens-num-clients}, our method consistently outperforms the baseline methods across all numbers of total client settings. 

\textbf{Hyperparameters $T_{\text{IG}}$ and $T_{\text{Agg}}$ for Aggregation.} Our default setting is $T_{\text{IG}}=$10 and $T_{\text{Agg}}=$10 . Varying $T_{\text{IG}}$ and $T_{\text{Agg}}$ show less than 1\% fluctuation in accuracy and consistently outperform other compared methods (compared in Table~\ref{tab:cifar100-results}). Nevertheless, we recommend selecting these hyperparameters based on the total number of clients and the client sampling rate.

\textbf{The size of local IG dataset $|D_{\text{IG}}|$.}
Our default setting uses $|D_{\text{IG}}| = 50$. As shown in Table~\ref{tab:sens-proxy-data-size}, reducing $|D_{\text{IG}}|$ to 5 or 10 results in only minor performance fluctuations, demonstrating that our method places minimal requirements on the size of the IG dataset. Since the IG scores are primarily used for model calibration, our method generalizes well even when clients have a limited amount of local data.

    


\begin{figure}[t]
\centering
\includegraphics[width=0.45\textwidth]{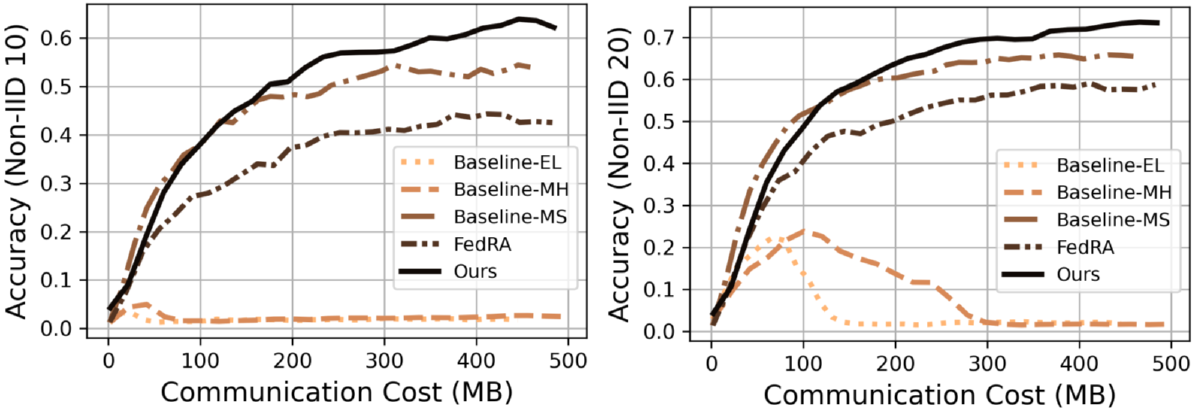}
\caption{Accuracy over communication cost on CIFAR-100 dataset (other methods not shown in the figure diverged).}
\label{fig:comm_eff}
\end{figure}

\begin{table}[t]
\centering
\caption{The impact of the size of local IG dataset $|D_{\text{IG}}|$. Experiments are conducted on CIFAR-100 under the Non-IID (10) setting.}
\scalebox{0.88}{
\begin{tabular}{cccc}
\toprule[1pt]
\textbf{Fixed Setting} & \textbf{5} & \textbf{10} & \textbf{50} \\
\hline
$T_{\text{Agg}}=10,\ T_{\text{IG}}=10$ & 62.78 & 62.60 & 62.93 \\
\bottomrule[1pt]
\end{tabular}
}
\label{tab:sens-proxy-data-size}
\end{table}

\subsection{Computation and Communication Efficiency}

We evaluate the computational and communication efficiency of our method by observing model accuracy over the cumulative training time and total communication cost during training, respectively. Compared to FedAvg, our method introduces additional computational cost every $T_{\text{IG}}$ rounds for optimizing the LoRA allocation, but reduces communication costs for low-resource clients by transmitting fewer parameters. 

\textbf{Computational Efficiency.} Figure~\ref{fig:comp_eff} presents the accuracy over training time for all methods on the Non-IID CIFAR-100 dataset. In this case, our method consistently achieves higher accuracy within the same training time (i.e., wall-clock time). Specifically, Baseline-EL and Baseline-MH diverge under the extreme Non-IID setting due to the lack of data from all the clients. Baseline-MS and FedRA do not incur additional computational costs compared to FedAvg, and therefore exhibit higher computational efficiency in the early stages (as shown in Figure~\ref{fig:comp_eff}, Non-IID 20). However, they suffer from accuracy degradation compared to our method due to their reliance on heuristic or stochastic LoRA allocation strategies. Although our method periodically incurs additional computational costs for LoRA allocation optimization, its accuracy gains lead to higher overall computational efficiency. 


\textbf{Communication Efficiency.} 
Figure~\ref{fig:comm_eff} presents the accuracy over communication cost for all methods on the Non-IID CIFAR-100 dataset.
We calculate both uplink and downlink communication costs per client during the training. Baseline-EL and Baseline-MH diverge but still incur a large amount of communication cost. FedRA and Baseline-MS both select a subset of LoRA modules to update locally, which also helps reduce the uplink communication costs for low-resource clients.  
Baseline-MS shows slightly higher computational efficiency in the early stages, as it enables more layers to be trained under the same memory constraint compared to FedRA and our method, leading to faster initial convergence.
However, this benefit diminishes over time, as it restricts training to only the last few layers, leaving the earlier layers untouched, which ultimately results in lower communication efficiency compared to our method. Overall, our method achieves a better trade-off between model accuracy and communication cost.

\section{Conclusion}\label{conclusion}
This paper presents Fed-pilot, a novel framework for optimizing LoRA allocation in federated fine-tuning while considering clients' memory capacities. We introduce the IG score to quantify each LoRA module's importance and design the optimization process that maximizes global model performance under heterogeneous memory constraints. Additionally, we develop a novel aggregation rule to handle diverse model updates under optimal LoRA allocation. Evaluations on five distinct FM fine-tuning tasks demonstrate the effectiveness and efficiency of Fed-pilot.

\bibliographystyle{ACM-Reference-Format}
\bibliography{main}


\begin{thebibliography}{37}


\ifx \showCODEN    \undefined \def \showCODEN     #1{\unskip}     \fi
\ifx \showDOI      \undefined \def \showDOI       #1{#1}\fi
\ifx \showISBNx    \undefined \def \showISBNx     #1{\unskip}     \fi
\ifx \showISBNxiii \undefined \def \showISBNxiii  #1{\unskip}     \fi
\ifx \showISSN     \undefined \def \showISSN      #1{\unskip}     \fi
\ifx \showLCCN     \undefined \def \showLCCN      #1{\unskip}     \fi
\ifx \shownote     \undefined \def \shownote      #1{#1}          \fi
\ifx \showarticletitle \undefined \def \showarticletitle #1{#1}   \fi
\ifx \showURL      \undefined \def \showURL       {\relax}        \fi
\providecommand\bibfield[2]{#2}
\providecommand\bibinfo[2]{#2}
\providecommand\natexlab[1]{#1}
\providecommand\showeprint[2][]{arXiv:#2}

\bibitem[Ardakani et~al\mbox{.}(2024)]%
        {ardakani2024slimfit}
\bibfield{author}{\bibinfo{person}{Arash Ardakani}, \bibinfo{person}{Altan Haan}, \bibinfo{person}{Shangyin Tan}, {et~al\mbox{.}}} \bibinfo{year}{2024}\natexlab{}.
\newblock \showarticletitle{SlimFit: Memory-Efficient Fine-Tuning of Transformer-based Models Using Training Dynamics}. In \bibinfo{booktitle}{\emph{Proceedings of the 2024 Conference of the NAACL: Human Language Technologies}}. \bibinfo{pages}{6218--6236}.
\newblock


\bibitem[Babakniya et~al\mbox{.}(2023)]%
        {babakniya2023slora}
\bibfield{author}{\bibinfo{person}{Sara Babakniya}, \bibinfo{person}{Ahmed~Roushdy Elkordy}, \bibinfo{person}{Yahya~H Ezzeldin}, {et~al\mbox{.}}} \bibinfo{year}{2023}\natexlab{}.
\newblock \showarticletitle{SLoRA: Federated Parameter Efficient Fine-Tuning of Language Models}. In \bibinfo{booktitle}{\emph{International Workshop in Conjunction with NeurIPS 2023}}.
\newblock


\bibitem[Bai et~al\mbox{.}(2024)]%
        {bai2024federated}
\bibfield{author}{\bibinfo{person}{Jiamu Bai}, \bibinfo{person}{Daoyuan Chen}, \bibinfo{person}{Bingchen Qian}, \bibinfo{person}{Liuyi Yao}, {and} \bibinfo{person}{Yaliang Li}.} \bibinfo{year}{2024}\natexlab{}.
\newblock \showarticletitle{Federated fine-tuning of large language models under heterogeneous tasks and client resources}. In \bibinfo{booktitle}{\emph{The Thirty-eighth Annual Conference on Neural Information Processing Systems}}.
\newblock


\bibitem[Chalkidis et~al\mbox{.}(2022)]%
        {chalkidis-etal-2021-lexglue}
\bibfield{author}{\bibinfo{person}{Ilias Chalkidis}, \bibinfo{person}{Abhik Jana}, \bibinfo{person}{Dirk Hartung}, {et~al\mbox{.}}} \bibinfo{year}{2022}\natexlab{}.
\newblock \showarticletitle{LexGLUE: A Benchmark Dataset for Legal Language Understanding in English}. In \bibinfo{booktitle}{\emph{Proceedings of the 60th Annual Meeting of the ACL}}. \bibinfo{address}{Dubln, Ireland}.
\newblock


\bibitem[Chen et~al\mbox{.}(2024)]%
        {chen2024data}
\bibfield{author}{\bibinfo{person}{Daoyuan Chen}, \bibinfo{person}{Yilun Huang}, \bibinfo{person}{Zhijian Ma}, \bibinfo{person}{Hesen Chen}, \bibinfo{person}{Xuchen Pan}, \bibinfo{person}{Ce Ge}, \bibinfo{person}{Dawei Gao}, \bibinfo{person}{Yuexiang Xie}, \bibinfo{person}{Zhaoyang Liu}, \bibinfo{person}{Jinyang Gao}, {et~al\mbox{.}}} \bibinfo{year}{2024}\natexlab{}.
\newblock \showarticletitle{Data-juicer: A one-stop data processing system for large language models}. In \bibinfo{booktitle}{\emph{Companion of the 2024 International Conference on Management of Data}}. \bibinfo{pages}{120--134}.
\newblock


\bibitem[Cho et~al\mbox{.}(2024)]%
        {cho2024heterogeneous}
\bibfield{author}{\bibinfo{person}{Yae~Jee Cho}, \bibinfo{person}{Luyang Liu}, \bibinfo{person}{Zheng Xu}, \bibinfo{person}{Aldi Fahrezi}, {and} \bibinfo{person}{Gauri Joshi}.} \bibinfo{year}{2024}\natexlab{}.
\newblock \showarticletitle{Heterogeneous lora for federated fine-tuning of on-device foundation models}. In \bibinfo{booktitle}{\emph{Proceedings of the 2024 Conference on Empirical Methods in Natural Language Processing}}. \bibinfo{pages}{12903--12913}.
\newblock


\bibitem[Conover et~al\mbox{.}(2023)]%
        {conover2023free}
\bibfield{author}{\bibinfo{person}{Mike Conover}, \bibinfo{person}{Matt Hayes}, \bibinfo{person}{Ankit Mathur}, \bibinfo{person}{Jianwei Xie}, \bibinfo{person}{Jun Wan}, \bibinfo{person}{Sam Shah}, \bibinfo{person}{Ali Ghodsi}, \bibinfo{person}{Patrick Wendell}, \bibinfo{person}{Matei Zaharia}, {and} \bibinfo{person}{Reynold Xin}.} \bibinfo{year}{2023}\natexlab{}.
\newblock \showarticletitle{Free dolly: Introducing the world’s first truly open instruction-tuned llm}.
\newblock \bibinfo{journal}{\emph{Company Blog of Databricks}} (\bibinfo{year}{2023}).
\newblock


\bibitem[Devlin et~al\mbox{.}(2019)]%
        {devlin2019bert}
\bibfield{author}{\bibinfo{person}{Jacob Devlin}, \bibinfo{person}{Ming-Wei Chang}, \bibinfo{person}{Kenton Lee}, {et~al\mbox{.}}} \bibinfo{year}{2019}\natexlab{}.
\newblock \showarticletitle{BERT: Pre-training of Deep Bidirectional Transformers for Language Understanding}. In \bibinfo{booktitle}{\emph{Proceedings of the 2019 Conference of the NAACL: Human Language Technologies, Volume 1}}.
\newblock


\bibitem[Dosovitskiy et~al\mbox{.}(2020)]%
        {dosovitskiy2020image}
\bibfield{author}{\bibinfo{person}{Alexey Dosovitskiy}, \bibinfo{person}{Lucas Beyer}, {et~al\mbox{.}}} \bibinfo{year}{2020}\natexlab{}.
\newblock \showarticletitle{An Image is Worth 16x16 Words: Transformers for Image Recognition at Scale}. In \bibinfo{booktitle}{\emph{ICLR}}.
\newblock


\bibitem[Gabri{\'e} et~al\mbox{.}(2018)]%
        {gabrie2018entropy}
\bibfield{author}{\bibinfo{person}{Marylou Gabri{\'e}}, \bibinfo{person}{Andre Manoel}, \bibinfo{person}{Cl{\'e}ment Luneau}, \bibinfo{person}{Nicolas Macris}, \bibinfo{person}{Florent Krzakala}, \bibinfo{person}{Lenka Zdeborov{\'a}}, {et~al\mbox{.}}} \bibinfo{year}{2018}\natexlab{}.
\newblock \showarticletitle{Entropy and mutual information in models of deep neural networks}.
\newblock \bibinfo{journal}{\emph{Advances in neural information processing systems}}  \bibinfo{volume}{31} (\bibinfo{year}{2018}).
\newblock


\bibitem[Gao et~al\mbox{.}(2025)]%
        {gao2025flowertune}
\bibfield{author}{\bibinfo{person}{Yan Gao}, \bibinfo{person}{Massimo~Roberto Scamarcia}, \bibinfo{person}{Javier Fernandez-Marques}, \bibinfo{person}{Mohammad Naseri}, \bibinfo{person}{Chong~Shen Ng}, \bibinfo{person}{Dimitris Stripelis}, \bibinfo{person}{Zexi Li}, \bibinfo{person}{Tao Shen}, \bibinfo{person}{Jiamu Bai}, \bibinfo{person}{Daoyuan Chen}, {et~al\mbox{.}}} \bibinfo{year}{2025}\natexlab{}.
\newblock \showarticletitle{FlowerTune: A Cross-Domain Benchmark for Federated Fine-Tuning of Large Language Models}.
\newblock \bibinfo{journal}{\emph{arXiv preprint arXiv:2506.02961}} (\bibinfo{year}{2025}).
\newblock


\bibitem[Guo et~al\mbox{.}(2024)]%
        {guo2024selective}
\bibfield{author}{\bibinfo{person}{Pengxin Guo}, \bibinfo{person}{Shuang Zeng}, \bibinfo{person}{Yanran Wang}, \bibinfo{person}{Huijie Fan}, \bibinfo{person}{Feifei Wang}, {and} \bibinfo{person}{Liangqiong Qu}.} \bibinfo{year}{2024}\natexlab{}.
\newblock \showarticletitle{Selective Aggregation for Low-Rank Adaptation in Federated Learning}.
\newblock \bibinfo{journal}{\emph{arXiv preprint arXiv:2410.01463}} (\bibinfo{year}{2024}).
\newblock


\bibitem[Hu et~al\mbox{.}(2021)]%
        {hu2021lora}
\bibfield{author}{\bibinfo{person}{Edward~J Hu}, \bibinfo{person}{Phillip Wallis}, \bibinfo{person}{Zeyuan Allen-Zhu}, {et~al\mbox{.}}} \bibinfo{year}{2021}\natexlab{}.
\newblock \showarticletitle{LoRA: Low-Rank Adaptation of Large Language Models}. In \bibinfo{booktitle}{\emph{ICLR}}.
\newblock


\bibitem[Kar et~al\mbox{.}(2023)]%
        {kar2023offloading}
\bibfield{author}{\bibinfo{person}{Binayak Kar}, \bibinfo{person}{Widhi Yahya}, \bibinfo{person}{Ying-Dar Lin}, {and} \bibinfo{person}{Asad Ali}.} \bibinfo{year}{2023}\natexlab{}.
\newblock \showarticletitle{Offloading using traditional optimization and machine learning in federated cloud--edge--fog systems: A survey}.
\newblock \bibinfo{journal}{\emph{IEEE Communications Surveys \& Tutorials}} \bibinfo{volume}{25}, \bibinfo{number}{2} (\bibinfo{year}{2023}), \bibinfo{pages}{1199--1226}.
\newblock


\bibitem[Kingma(2014)]%
        {kingma2014adam}
\bibfield{author}{\bibinfo{person}{Diederik~P Kingma}.} \bibinfo{year}{2014}\natexlab{}.
\newblock \showarticletitle{Adam: A method for stochastic optimization}.
\newblock \bibinfo{journal}{\emph{arXiv preprint arXiv:1412.6980}} (\bibinfo{year}{2014}).
\newblock


\bibitem[Krizhevsky et~al\mbox{.}(2009)]%
        {krizhevsky2009learning}
\bibfield{author}{\bibinfo{person}{Alex Krizhevsky}, \bibinfo{person}{Geoffrey Hinton}, {et~al\mbox{.}}} \bibinfo{year}{2009}\natexlab{}.
\newblock \showarticletitle{Learning multiple layers of features from tiny images}. In \bibinfo{booktitle}{\emph{Toronto, ON, Canada}}.
\newblock


\bibitem[Kullback(1997)]%
        {kullback1997information}
\bibfield{author}{\bibinfo{person}{Solomon Kullback}.} \bibinfo{year}{1997}\natexlab{}.
\newblock \bibinfo{booktitle}{\emph{Information theory and statistics}}.
\newblock \bibinfo{publisher}{Courier Corporation}.
\newblock


\bibitem[Lester et~al\mbox{.}(2021)]%
        {lester2021power}
\bibfield{author}{\bibinfo{person}{Brian Lester}, \bibinfo{person}{Rami Al-Rfou}, {and} \bibinfo{person}{Noah Constant}.} \bibinfo{year}{2021}\natexlab{}.
\newblock \showarticletitle{The Power of Scale for Parameter-Efficient Prompt Tuning}. In \bibinfo{booktitle}{\emph{Proceedings of the 2021 Conference on EMNLP}}. \bibinfo{pages}{3045--3059}.
\newblock


\bibitem[Liu et~al\mbox{.}(2024)]%
        {liu2024fisher}
\bibfield{author}{\bibinfo{person}{Ji Liu}, \bibinfo{person}{Jiaxiang Ren}, \bibinfo{person}{Ruoming Jin}, \bibinfo{person}{Zijie Zhang}, \bibinfo{person}{Yang Zhou}, \bibinfo{person}{Patrick Valduriez}, {and} \bibinfo{person}{Dejing Dou}.} \bibinfo{year}{2024}\natexlab{}.
\newblock \showarticletitle{Fisher Information-based Efficient Curriculum Federated Learning with Large Language Models}. In \bibinfo{booktitle}{\emph{Proceedings of the 2024 Conference on Empirical Methods in Natural Language Processing}}. \bibinfo{pages}{10497--10523}.
\newblock


\bibitem[McMahan et~al\mbox{.}(2017)]%
        {mcmahan2017communication}
\bibfield{author}{\bibinfo{person}{Brendan McMahan}, \bibinfo{person}{Eider Moore}, \bibinfo{person}{Daniel Ramage}, {et~al\mbox{.}}} \bibinfo{year}{2017}\natexlab{}.
\newblock \showarticletitle{Communication-efficient learning of deep networks from decentralized data}. In \bibinfo{booktitle}{\emph{Artificial intelligence and statistics}}. PMLR.
\newblock


\bibitem[Pan et~al\mbox{.}(2024)]%
        {pan2024lisa}
\bibfield{author}{\bibinfo{person}{Rui Pan}, \bibinfo{person}{Xiang Liu}, \bibinfo{person}{Shizhe Diao}, \bibinfo{person}{Renjie Pi}, \bibinfo{person}{Jipeng Zhang}, \bibinfo{person}{Chi Han}, {and} \bibinfo{person}{Tong Zhang}.} \bibinfo{year}{2024}\natexlab{}.
\newblock \showarticletitle{LISA: layerwise importance sampling for memory-efficient large language model fine-tuning}.
\newblock \bibinfo{journal}{\emph{Advances in Neural Information Processing Systems}}  \bibinfo{volume}{37} (\bibinfo{year}{2024}), \bibinfo{pages}{57018--57049}.
\newblock


\bibitem[Peng et~al\mbox{.}(2019)]%
        {peng2019moment}
\bibfield{author}{\bibinfo{person}{Xingchao Peng}, \bibinfo{person}{Qinxun Bai}, \bibinfo{person}{Xide Xia}, {et~al\mbox{.}}} \bibinfo{year}{2019}\natexlab{}.
\newblock \showarticletitle{Moment matching for multi-source domain adaptation}. In \bibinfo{booktitle}{\emph{Proceedings of the ICCV}}.
\newblock


\bibitem[Rasley et~al\mbox{.}(2020)]%
        {rasley2020deepspeed}
\bibfield{author}{\bibinfo{person}{Jeff Rasley}, \bibinfo{person}{Samyam Rajbhandari}, \bibinfo{person}{Olatunji Ruwase}, {and} \bibinfo{person}{Yuxiong He}.} \bibinfo{year}{2020}\natexlab{}.
\newblock \showarticletitle{Deepspeed: System optimizations enable training deep learning models with over 100 billion parameters}. In \bibinfo{booktitle}{\emph{Proceedings of the 26th ACM SIGKDD international conference on knowledge discovery \& data mining}}. \bibinfo{pages}{3505--3506}.
\newblock


\bibitem[Su et~al\mbox{.}(2023)]%
        {su2023fedra}
\bibfield{author}{\bibinfo{person}{Shangchao Su}, \bibinfo{person}{Bin Li}, {and} \bibinfo{person}{Xiangyang Xue}.} \bibinfo{year}{2023}\natexlab{}.
\newblock \showarticletitle{FedRA: A Random Allocation Strategy for Federated Tuning to Unleash the Power of Heterogeneous Clients}.
\newblock \bibinfo{journal}{\emph{arXiv preprint arXiv:2311.11227}} (\bibinfo{year}{2023}).
\newblock


\bibitem[Sun et~al\mbox{.}(2024)]%
        {sun2024improving}
\bibfield{author}{\bibinfo{person}{Youbang Sun}, \bibinfo{person}{Zitao Li}, \bibinfo{person}{Yaliang Li}, {and} \bibinfo{person}{Bolin Ding}.} \bibinfo{year}{2024}\natexlab{}.
\newblock \showarticletitle{Improving Lo{RA} in Privacy-preserving Federated Learning}. In \bibinfo{booktitle}{\emph{The Twelfth International Conference on Learning Representations}}.
\newblock
\urldef\tempurl%
\url{https://openreview.net/forum?id=NLPzL6HWNl}
\showURL{%
\tempurl}


\bibitem[Taori et~al\mbox{.}(2023)]%
        {taori2023stanford}
\bibfield{author}{\bibinfo{person}{Rohan Taori}, \bibinfo{person}{Ishaan Gulrajani}, \bibinfo{person}{Tianyi Zhang}, \bibinfo{person}{Yann Dubois}, \bibinfo{person}{Xuechen Li}, \bibinfo{person}{Carlos Guestrin}, \bibinfo{person}{Percy Liang}, {and} \bibinfo{person}{Tatsunori~B Hashimoto}.} \bibinfo{year}{2023}\natexlab{}.
\newblock \bibinfo{title}{Stanford alpaca: An instruction-following llama model}.
\newblock
\newblock


\bibitem[Touvron et~al\mbox{.}(2023)]%
        {touvron2023llama}
\bibfield{author}{\bibinfo{person}{Hugo Touvron}, \bibinfo{person}{Thibaut Lavril}, \bibinfo{person}{Gautier Izacard}, \bibinfo{person}{Xavier Martinet}, \bibinfo{person}{Marie-Anne Lachaux}, \bibinfo{person}{Timoth{\'e}e Lacroix}, \bibinfo{person}{Baptiste Rozi{\`e}re}, \bibinfo{person}{Naman Goyal}, \bibinfo{person}{Eric Hambro}, \bibinfo{person}{Faisal Azhar}, {et~al\mbox{.}}} \bibinfo{year}{2023}\natexlab{}.
\newblock \showarticletitle{Llama: Open and efficient foundation language models}.
\newblock \bibinfo{journal}{\emph{arXiv preprint arXiv:2302.13971}} (\bibinfo{year}{2023}).
\newblock


\bibitem[Wang et~al\mbox{.}(2022)]%
        {wang2022super}
\bibfield{author}{\bibinfo{person}{Yizhong Wang}, \bibinfo{person}{Swaroop Mishra}, \bibinfo{person}{Pegah Alipoormolabashi}, \bibinfo{person}{Yeganeh Kordi}, \bibinfo{person}{Amirreza Mirzaei}, \bibinfo{person}{Atharva Naik}, \bibinfo{person}{Arjun Ashok}, \bibinfo{person}{Arut~Selvan Dhanasekaran}, \bibinfo{person}{Anjana Arunkumar}, \bibinfo{person}{David Stap}, {et~al\mbox{.}}} \bibinfo{year}{2022}\natexlab{}.
\newblock \showarticletitle{Super-NaturalInstructions: Generalization via Declarative Instructions on 1600+ NLP Tasks}. In \bibinfo{booktitle}{\emph{Proceedings of the 2022 Conference on Empirical Methods in Natural Language Processing}}. \bibinfo{pages}{5085--5109}.
\newblock


\bibitem[Wang et~al\mbox{.}(2024)]%
        {wangflora}
\bibfield{author}{\bibinfo{person}{Ziyao Wang}, \bibinfo{person}{Zheyu Shen}, \bibinfo{person}{Yexiao He}, \bibinfo{person}{Guoheng Sun}, \bibinfo{person}{Hongyi Wang}, \bibinfo{person}{Lingjuan Lyu}, {and} \bibinfo{person}{Ang Li}.} \bibinfo{year}{2024}\natexlab{}.
\newblock \showarticletitle{FLoRA: Federated Fine-Tuning Large Language Models with Heterogeneous Low-Rank Adaptations}. In \bibinfo{booktitle}{\emph{The Thirty-eighth Annual Conference on Neural Information Processing Systems}}.
\newblock


\bibitem[Wu et~al\mbox{.}(2024)]%
        {wu2024fedfmsl}
\bibfield{author}{\bibinfo{person}{Panlong Wu}, \bibinfo{person}{Kangshuo Li}, \bibinfo{person}{Ting Wang}, \bibinfo{person}{Yanjie Dong}, \bibinfo{person}{Victor~CM Leung}, {and} \bibinfo{person}{Fangxin Wang}.} \bibinfo{year}{2024}\natexlab{}.
\newblock \showarticletitle{FedFMSL: Federated Learning of Foundations Models With Sparsely Activated LoRA}.
\newblock \bibinfo{journal}{\emph{IEEE Transactions on Mobile Computing}} (\bibinfo{year}{2024}).
\newblock


\bibitem[Xu et~al\mbox{.}(2024)]%
        {xu2024fwdllm}
\bibfield{author}{\bibinfo{person}{Mengwei Xu}, \bibinfo{person}{Dongqi Cai}, \bibinfo{person}{Yaozong Wu}, \bibinfo{person}{Xiang Li}, {and} \bibinfo{person}{Shangguang Wang}.} \bibinfo{year}{2024}\natexlab{}.
\newblock \showarticletitle{FwdLLM: Efficient federated finetuning of large language models with perturbed inferences}. In \bibinfo{booktitle}{\emph{USENIX ATC}}.
\newblock


\bibitem[Yao et~al\mbox{.}(2024)]%
        {yao2024layer}
\bibfield{author}{\bibinfo{person}{Kai Yao}, \bibinfo{person}{Penglei Gao}, \bibinfo{person}{Lichun Li}, \bibinfo{person}{Yuan Zhao}, \bibinfo{person}{Xiaofeng Wang}, \bibinfo{person}{Wei Wang}, {and} \bibinfo{person}{Jianke Zhu}.} \bibinfo{year}{2024}\natexlab{}.
\newblock \showarticletitle{Layer-wise Importance Matters: Less Memory for Better Performance in Parameter-efficient Fine-tuning of Large Language Models}. In \bibinfo{booktitle}{\emph{Findings of the Association for Computational Linguistics: EMNLP 2024}}. \bibinfo{pages}{1977--1992}.
\newblock


\bibitem[Zhang et~al\mbox{.}(2023)]%
        {zhang2023towards}
\bibfield{author}{\bibinfo{person}{Jianyi Zhang}, \bibinfo{person}{Saeed Vahidian}, \bibinfo{person}{Martin Kuo}, {et~al\mbox{.}}} \bibinfo{year}{2023}\natexlab{}.
\newblock \showarticletitle{Towards Building the FederatedGPT: Federated Instruction Tuning}. In \bibinfo{booktitle}{\emph{International Workshop in Conjunction with NeurIPS 2023}}.
\newblock


\bibitem[Zhang et~al\mbox{.}(2022)]%
        {zhang2022opt}
\bibfield{author}{\bibinfo{person}{Susan Zhang}, \bibinfo{person}{Stephen Roller}, \bibinfo{person}{Naman Goyal}, \bibinfo{person}{Mikel Artetxe}, \bibinfo{person}{Moya Chen}, \bibinfo{person}{Shuohui Chen}, \bibinfo{person}{Christopher Dewan}, \bibinfo{person}{Mona Diab}, \bibinfo{person}{Xian Li}, \bibinfo{person}{Xi~Victoria Lin}, {et~al\mbox{.}}} \bibinfo{year}{2022}\natexlab{}.
\newblock \showarticletitle{Opt: Open pre-trained transformer language models}.
\newblock \bibinfo{journal}{\emph{arXiv preprint arXiv:2205.01068}} (\bibinfo{year}{2022}).
\newblock


\bibitem[Zhang et~al\mbox{.}(2025)]%
        {zhang2025fed}
\bibfield{author}{\bibinfo{person}{Zikai Zhang}, \bibinfo{person}{Ping Liu}, \bibinfo{person}{Jiahao Xu}, {and} \bibinfo{person}{Rui Hu}.} \bibinfo{year}{2025}\natexlab{}.
\newblock \showarticletitle{Fed-HeLLo: Efficient Federated Foundation Model Fine-Tuning with Heterogeneous LoRA Allocation}.
\newblock \bibinfo{journal}{\emph{arXiv preprint arXiv:2506.12213}} (\bibinfo{year}{2025}).
\newblock


\bibitem[Zhu et~al\mbox{.}(2023)]%
        {zhu2023lift}
\bibfield{author}{\bibinfo{person}{Ligeng Zhu}, \bibinfo{person}{Lanxiang Hu}, \bibinfo{person}{Ji Lin}, {and} \bibinfo{person}{Song Han}.} \bibinfo{year}{2023}\natexlab{}.
\newblock \showarticletitle{Lift: Efficient layer-wise fine-tuning for large model models}.
\newblock  (\bibinfo{year}{2023}).
\newblock


\bibitem[Zhu et~al\mbox{.}(2021)]%
        {zhu2021data}
\bibfield{author}{\bibinfo{person}{Zhuangdi Zhu}, \bibinfo{person}{Junyuan Hong}, {and} \bibinfo{person}{Jiayu Zhou}.} \bibinfo{year}{2021}\natexlab{}.
\newblock \showarticletitle{Data-free knowledge distillation for heterogeneous federated learning}. In \bibinfo{booktitle}{\emph{International conference on machine learning}}. PMLR, \bibinfo{pages}{12878--12889}.
\newblock


\end{thebibliography}

\newpage
\appendix
\onecolumn
\setcounter{page}{1}

\section*{Supplement}

We provide more discussions and experiments of this work in the supplement and organize it as follows:

\begin{itemize}
    \item Supplement~\ref{sec:appendix-gpu-memory-consumption}: Additional details on memory consumption in LoRA-based FM fine-tuning.
        \begin{itemize}
            \item In Subsection~\ref{subsec:appendix-dynamic-activation}, we provide an example of how to calculate activations with only a frozen linear layer.
            \item In Subsection~\ref{subsec:appendix-static-activation}, we provide an example of how to calculate activations with a frozen non-linear layer.
            \item In Subsection~\ref{subsec:appendix-dynamic-static-activation}, we provide details on the type of activation in the ViT model.
            \item In Subsection~\ref{subsec:appendix-memory-footprint-example}, we provide a code of how to calculate memory consumption for a given LoRA allocation map with ViT-base.
        \end{itemize}

    
    \item Supplement~\ref{sec:appendix-settings}: We provide detailed information on the experimental settings, including the models, FL settings, GPU memory capacity configurations for each client and model level, the resources and infrastructure used for the experiments, and a detailed explanation of the experiments shown in Figure~\ref{fig:motivation}.
        \begin{itemize}
            \item In Subsection~\ref{subsec:appendix-settings-flsettings}, we outline the FL-related settings and hyperparameters.
            \item In Subsection~\ref{subsec:appendix-capacity-settings}, we describe the capacity chooses for each model.
            \item In Subsection~\ref{subsec:appendix-motivation-settings}, we provide a detailed explanation of the experimental settings used for Figure~\ref{fig:motivation}.
        \end{itemize}
   
    \item Supplement~\ref{sec:appendix-dataset}: Details on the preprocessing of all datasets are provided.

    \item Supplement~\ref{sec:appendix-results-visualization}: Additional results are presented in tables.
    
    \item Supplement~\ref{sec:appendix-comparative_study}: A comparative study on state-of-the-art methods is provided.

    \item Supplement~\ref{sec:appendix-limitation}: The limitations of our work are discussed.
   
    \item Supplement~\ref{sec:appendix-futurework}: Directions for future work are outlined.

\end{itemize}



\section{GPU Memory Consumption of Model Fine-Tuning with LoRA}\label{sec:appendix-gpu-memory-consumption}

    \subsection{Backpropagation for Linear Layers: Saving Dynamic Activation using Freezing}\label{subsec:appendix-dynamic-activation}
    
    We provide a simple example~\cite{ardakani2024slimfit} demonstrating how gradients are backpropagated to the first layer of a neural network while its middle layer is frozen. To this end, let us perform backpropagation using a 3-layer neural network as an example. The architecture of this network is described as follows,
    
    \begin{align}
        \mathbf{y}_1 &= \mathbf{x}\mathbf{W}_1 + \mathbf{b}_1, \\
        \mathbf{y}_2 &= \mathbf{y}_1\mathbf{W}_2 + \mathbf{b}_2, \\
        \mathbf{y}_3 &= \mathbf{y}_2\mathbf{W}_3 + \mathbf{b}_3.
    \end{align}
    
    where $\{\mathbf{W}_1, \mathbf{W}_2, \mathbf{W}_3$\} and $\{\mathbf{b}_1, \mathbf{b}_2, \mathbf{b}_3\}$ are the weights and biases of the network, respectively. In this example, $\mathbf{x}$, $\mathbf{y}_1$, and $\mathbf{y}_2$ are input to the first layer, the second layer, and the third layer, respectively. 
    The backpropagation equations for the loss $\mathcal{L}$ using the chain rule can be presented as follows,
    
    \begin{align}
        \frac{\partial \mathcal{L}}{\partial \mathbf{W}_3} &= \frac{\partial \mathcal{L}}{\partial \mathbf{y}_3} \frac{\partial \mathbf{y}_3}{\partial \mathbf{W}_3} = \frac{\partial \mathcal{L}}{\partial \mathbf{y}_3} \mathbf{y}_2, \\
        \frac{\partial \mathcal{L}}{\partial \mathbf{b}_3} &= \frac{\partial \mathcal{L}}{\partial \mathbf{y}_3} \frac{\partial \mathbf{y}_3}{\partial \mathbf{b}_3} = \frac{\partial \mathcal{L}}{\partial \mathbf{y}_3}, \\
        \frac{\partial \mathcal{L}}{\partial \mathbf{W}_2} &= \frac{\partial \mathcal{L}}{\partial \mathbf{y}_3} \frac{\partial \mathbf{y}_3}{\partial \mathbf{y}_2} \frac{\partial \mathbf{y}_2}{\partial \mathbf{W}_2} = \frac{\partial \mathcal{L}}{\partial \mathbf{y}_3} \mathbf{W}_3^\top \mathbf{y}_1, \\
        \frac{\partial \mathcal{L}}{\partial \mathbf{b}_2} &= \frac{\partial \mathcal{L}}{\partial \mathbf{y}_3} \frac{\partial \mathbf{y}_3}{\partial \mathbf{y}_2} \frac{\partial \mathbf{y}_2}{\partial \mathbf{b}_2} = \frac{\partial \mathcal{L}}{\partial \mathbf{y}_3} \mathbf{W}_3^\top, \\
        \frac{\partial \mathcal{L}}{\partial \mathbf{W}_1} &= \frac{\partial \mathcal{L}}{\partial \mathbf{y}_3} \frac{\partial \mathbf{y}_3}{\partial \mathbf{y}_2} \frac{\partial \mathbf{y}_2}{\partial \mathbf{y}_1} \frac{\partial \mathbf{y}_1}{\partial \mathbf{W}_1} = \frac{\partial \mathcal{L}}{\partial \mathbf{y}_3} \mathbf{W}_3^\top \mathbf{W}_2^\top \mathbf{x}, \\
        \frac{\partial \mathcal{L}}{\partial \mathbf{b}_1} &= \frac{\partial \mathcal{L}}{\partial \mathbf{y}_3} \frac{\partial \mathbf{y}_3}{\partial \mathbf{y}_2} \frac{\partial \mathbf{y}_2}{\partial \mathbf{y}_1} \frac{\partial \mathbf{y}_1}{\partial \mathbf{b}_1} = \frac{\partial \mathcal{L}}{\partial \mathbf{y}_3} \mathbf{W}_3^\top \mathbf{W}_2^\top.
    \end{align}

    where $\partial$ denotes the partial derivative, and $\frac{\partial \mathcal{L}}{\partial \mathbf{W}_3}$ is obtained by computing the loss.
    
    \textbf{Training all three layers.} To update all the network weights (i.e., $\mathbf{W}_1, \mathbf{W}_2, \mathbf{W}_3$), we need to store $\mathbf{x}$, $\mathbf{y}_1$, and $\mathbf{y}_2$ during forward computations, as they are required in the gradient calculations.
    
    \textbf{Training two layers while freezing the middle layer.} In the case of freezing the middle layer, it is unnecessary to compute the gradient update for $\mathbf{W}_2$:
    
    \begin{equation}
        \frac{\partial \mathcal{L}}{\partial \mathbf{W}_2} = \frac{\partial \mathcal{L}}{\partial \mathbf{y}_3} \mathbf{W}_3^\top \mathbf{y}_1.
    \end{equation}
    
    Therefore, there is no need to store $\mathbf{y}_1$ during the forward pass. However, discarding $\mathbf{y}_1$ does not affect the backward computations of the first layer, since the updates for $\mathbf{W}_1$ and $\mathbf{b}_1$ are independent of $\mathbf{y}_1$:
    
    \begin{align}
        \frac{\partial \mathcal{L}}{\partial \mathbf{W}_1} &= \frac{\partial \mathcal{L}}{\partial \mathbf{y}_3} \mathbf{W}_3^\top \mathbf{W}_2^\top \mathbf{x}, \\
        \frac{\partial \mathcal{L}}{\partial \mathbf{b}_1} &= \frac{\partial \mathcal{L}}{\partial \mathbf{y}_3} \mathbf{W}_3^\top \mathbf{W}_2^\top \mathbf{1}.
    \end{align}
    
    This example illustrates how freezing an intermediate linear layer affects the backpropagation process. For a liner layer, activation can be discarded if the layer is frozen.

    \subsection{Backpropagation for Non-Linear Layers: Static Activation Remains using Freezing
    }\label{subsec:appendix-static-activation}

    Here, we demonstrate how non-linear transformations (such as Softmax, GELU, and LayerNorm) impact backpropagation in a 3-layer neural network. By extending the example in~\cite{ardakani2024slimfit}, the architecture of the network is defined as follows:

    
    \begin{align}
        \mathbf{y}_1 &= \mathbf{x}\mathbf{W}_1 + \mathbf{b}_1, \\
        \mathbf{y}_2 &= \sigma(\mathbf{y}_1\mathbf{W}_2 + \mathbf{b}_2), \\
        \mathbf{y}_3 &= \mathbf{y}_2\mathbf{W}_3 + \mathbf{b}_3.
    \end{align}
    
    where $\{\mathbf{W}_1, \mathbf{W}_2, \mathbf{W}_3\}$ and $\{\mathbf{b}_1, \mathbf{b}_2, \mathbf{b}_3\}$ are the weights and biases of the network, respectively. In this example, $\mathbf{x}$, $\mathbf{y}_1$, and $\mathbf{y}_2$ are input to the first layer, the second layer, and the third layer, respectively. The non-linear function $\sigma(\cdot)$ introduces non-linearity in the second layer. The backpropagation equations for the loss $\mathcal{L}$ using the chain rule can be presented as follows,
    
    \begin{align}
        \frac{\partial \mathcal{L}}{\partial \mathbf{W}_3} &= \frac{\partial \mathcal{L}}{\partial \mathbf{y}_3} \frac{\partial \mathbf{y}_3}{\partial \mathbf{W}_3} = \frac{\partial \mathcal{L}}{\partial \mathbf{y}_3} \mathbf{y}_2, \\
        \frac{\partial \mathcal{L}}{\partial \mathbf{b}_3} &= \frac{\partial \mathcal{L}}{\partial \mathbf{y}_3} \frac{\partial \mathbf{y}_3}{\partial \mathbf{b}_3} = \frac{\partial \mathcal{L}}{\partial \mathbf{y}_3}, \\
        \frac{\partial \mathcal{L}}{\partial \mathbf{W}_2} &= \frac{\partial \mathcal{L}}{\partial \mathbf{y}_3} \frac{\partial \mathbf{y}_3}{\partial \mathbf{y}_2} \frac{\partial \mathbf{y}_2}{\partial \mathbf{W}_2} = \frac{\partial \mathcal{L}}{\partial \mathbf{y}_3} \mathbf{W}_3^\top \sigma'(\mathbf{y}_1\mathbf{W}_2 + \mathbf{b}_2) \mathbf{y}_1, \\
        \frac{\partial \mathcal{L}}{\partial \mathbf{b}_2} &= \frac{\partial \mathcal{L}}{\partial \mathbf{y}_3} \frac{\partial \mathbf{y}_3}{\partial \mathbf{y}_2} \frac{\partial \mathbf{y}_2}{\partial \mathbf{b}_2} = \frac{\partial \mathcal{L}}{\partial \mathbf{y}_3} \mathbf{W}_3^\top \sigma'(\mathbf{y}_1\mathbf{W}_2 + \mathbf{b}_2), \\
        \frac{\partial \mathcal{L}}{\partial \mathbf{W}_1} &= \frac{\partial \mathcal{L}}{\partial \mathbf{y}_3} \frac{\partial \mathcal{L}}{\partial \mathbf{y}_3}{\partial \mathbf{y}_2} \frac{\partial \mathbf{y}_2}{\partial \mathbf{y}_1} \frac{\partial \mathbf{y}_1}{\partial \mathbf{W}_1} = \frac{\partial \mathcal{L}}{\partial \mathbf{y}_3} \mathbf{W}_3^\top \sigma'(\mathbf{y}_1\mathbf{W}_2 + \mathbf{b}_2) \mathbf{W}_2^\top \mathbf{x}, \\
        \frac{\partial \mathcal{L}}{\partial \mathbf{b}_1} &= \frac{\partial \mathcal{L}}{\partial \mathbf{y}_3} \frac{\partial \mathcal{L}}{\partial \mathbf{y}_3}{\partial \mathbf{y}_2} \frac{\partial \mathbf{y}_2}{\partial \mathbf{y}_1} \frac{\partial \mathbf{y}_1}{\partial \mathbf{b}_1} = \frac{\partial \mathcal{L}}{\partial \mathbf{y}_3} \mathbf{W}_3^\top \sigma'(\mathbf{y}_1\mathbf{W}_2 + \mathbf{b}_2) \mathbf{W}_2^\top.
    \end{align}
    Here, $\sigma^{\prime}(\mathbf{y}_1\mathbf{W}_2)$ represents the derivative of the non-linear function $\sigma(\cdot)$ applied to the linear transformation $\mathbf{y}_1\mathbf{W}_2 + \mathbf{b}_2$.
    
    \textbf{Training all three layers.} To update all the network weights (i.e., $\mathbf{W}_1, \mathbf{W}_2, \mathbf{W}_3$), we need to store $\mathbf{x}$, $\mathbf{y}_1$, and $\mathbf{y}_2$ during forward computations, as they are required in the gradient calculations.
    
    \textbf{Training two layers while freezing the middle layer.} In the case of freezing the middle layer, it is unnecessary to compute the gradient update for $\mathbf{W}_2$:

    \begin{equation}
        \frac{\partial \mathcal{L}}{\partial \mathbf{W}_2} = \frac{\partial \mathcal{L}}{\partial \mathbf{y}_3} \mathbf{W}_3^\top \sigma'(\mathbf{y}_1\mathbf{W}_2 + \mathbf{b}_2) \mathbf{y}_1.
    \end{equation}

    However, unlike the linear case, it is \textbf{necessary} to store $\mathbf{y}_1$ during the forward pass because the gradients for $\mathbf{w}_1$ and $\mathbf{b}_1$ depends on $\sigma'(\mathbf{y}_1\mathbf{W}_2 + \mathbf{b}_2)$. Discarding $\mathbf{y}_1$ would prevent the correct computation of gradients for $\mathbf{W}_1$ and $\mathbf{b}_1$, as their updates explicitly involve $\mathbf{y}_1$:

    \begin{align}
        \frac{\partial \mathcal{L}}{\partial \mathbf{W}_1} &= \frac{\partial \mathcal{L}}{\partial \mathbf{y}_3} \mathbf{W}_3^\top \sigma'(\mathbf{y}_1\mathbf{W}_2 + \mathbf{b}_2) \mathbf{W}_2^\top \mathbf{x}, \\
        \frac{\partial \mathcal{L}}{\partial \mathbf{b}_1} &= \frac{\partial \mathcal{L}}{\partial \mathbf{y}_3} \mathbf{W}_3^\top \sigma'(\mathbf{y}_1\mathbf{W}_2 + \mathbf{b}_2) \mathbf{W}_2^\top.
    \end{align}

    This example illustrates how freezing an intermediate non-linear layer affects the backpropagation process. Unlike the linear case, where activations can be discarded when a layer is frozen, a non-linear layer requires storing its activations due to its role in computing the gradients of earlier layers.

    \begin{figure}[t]
    \centering
    \includegraphics[width=0.45\textwidth]{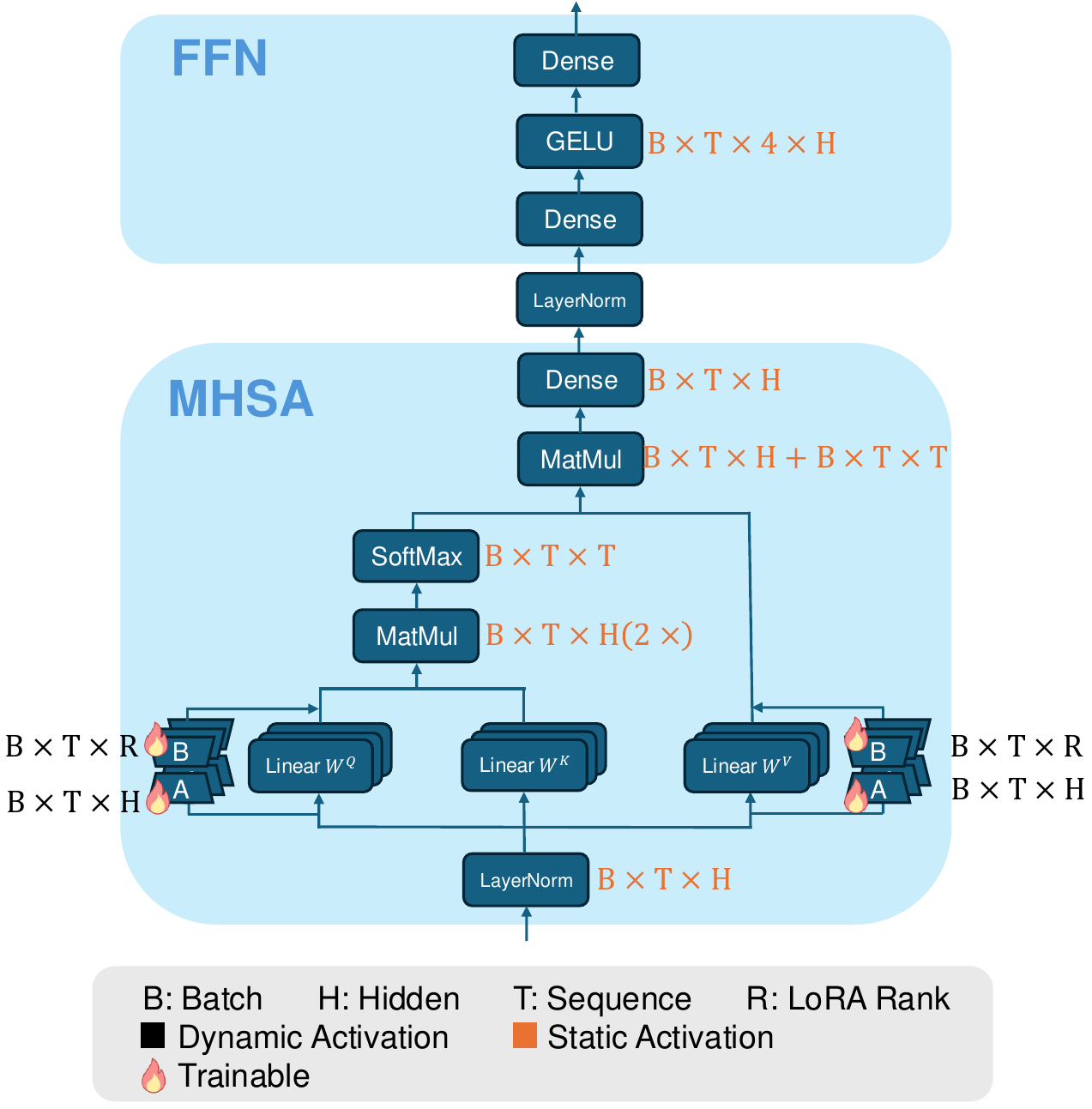}
    \caption{Illustration of activation memory usage in a transformer block for the ViT-base model (the rest layers are frozen if not specified.}
    \label{fig:memorycost}
    \end{figure}

    \subsection{Dynamic and Static Activations: An Example for ViT-base with LoRA module}\label{subsec:appendix-dynamic-static-activation}

    \begin{table}[t]
    \centering
    \caption{Details of transformer block of for ViT-base, including the internal layer types, descriptions, activation sizes, and type of activations~\cite{ardakani2024slimfit}.}
    \caption{Memory consumption estimates for training different LoRA modules.}
    \scalebox{1.2}{
    \begin{tabular}{c|c|c|c|c}
    \toprule[1pt]
    \textbf{Module} & \textbf{Type of Layer} & \textbf{Description} & \textbf{\# Activations}  & \textbf{Type of Activations}\\ \hline
    \multirow{8}{*}{MHSA} 
        & Dense     & attention.query    & $B \times T \times H$ & Dynamic \\ 
        & Dense     & attention.query.lora.B    & $B \times T \times R$ & Dynamic \\ 
        & Dense     & attention.value    & $B \times T \times H$ & Dynamic\\ 
        & Dense     & attention.value.lora.B    & $B \times T \times R$ & Dynamic\\ 
        & MatMul    & -                 & $B \times T \times H (2\times)$ &\textbf{Static}\\ 
        & Softmax   & -                 & $B \times T \times T$ & \textbf{Static}\\ 
        & MatMul    & -                 & $B \times T \times H$ \& $B \times T \times T$ &\textbf{Static}\\ 
        & LayerNorm & attention.output   & $B \times T \times H$ &\textbf{Static} \\ \hline
    \multirow{2}{*}{FFN} 
        & GELU      & -                 & $B \times T \times 4H$ &\textbf{Static}\\ 
        & LayerNorm & output             & $B \times T \times H$ &\textbf{Static}\\ \bottomrule[1pt]
    \end{tabular}
    }
    \label{tab:activations}
    \end{table}

    In the previous sections, we defined dynamic and static activations. Here, we provide a real example using the ViT-base~\cite{dosovitskiy2020image} model, which consists of 12 transformer blocks with LoRA modules added to the attention layers. Figure~\ref{fig:memorycost} illustrates the activation memory usage in a transformer block of ViT-base, where only the LoRA modules are trainable.

    Among different operations, GELU, MatMul, Softmax, and LayerNorm involve static activations, as summarized in Table~\ref{tab:activations}. Notably, MatMul and Softmax share the same activations—Softmax stores its output during the forward pass for reuse in backward computations, while MatMul requires its input for backpropagation. Since Softmax’s output serves as MatMul’s input during the forward pass, both operations rely on the same activations.

    \subsection{Code for Calculating GPU Memory Consumption}\label{subsec:appendix-memory-footprint-example}

\definecolor{codegreen}{rgb}{0,0.6,0}
\definecolor{codegray}{rgb}{0.5,0.5,0.5}
\definecolor{codepurple}{rgb}{0.58,0,0.82}
\definecolor{backcolour}{rgb}{0.95,0.95,0.92}

\lstdefinestyle{mystyle}{
    backgroundcolor=\color{backcolour},   
    commentstyle=\color{codegreen},
    keywordstyle=\color{blue},
    numberstyle=\tiny\color{codegray},
    stringstyle=\color{codepurple},
    basicstyle=\ttfamily\footnotesize,
    breakatwhitespace=false,         
    breaklines=true,                 
    captionpos=b,                    
    keepspaces=true,                 
    numbers=left,                    
    numbersep=5pt,                  
    showspaces=false,                
    showstringspaces=false,
    showtabs=false,                  
    tabsize=4
}

\lstset{style=mystyle}

We provided a script that estimates the memory consumption for the ViT-base model when using LoRA-based fine-tuning. The calculations include static and dynamic activation memory, parameter memory, optimizer memory, and GPU context. Three configurations are evaluated: training the full 12 LoRA modules, the first 6 LoRA modules, and the last 6 LoRA modules.

\begin{lstlisting}[language=Python, caption=Memory Consumption Estimation for ViT with LoRA]
from transformers import AutoModelForImageClassification
from peft import LoraConfig, get_peft_model
        
def get_activation(model, rank, sequence_size, hidden_size, batch_size):
    element_size = 4  # For float32
    static_activation_perL = (9 * batch_size * sequence_size * hidden_size + 
                              2 * batch_size * sequence_size * sequence_size) * element_size / (10**6)
    dynamic_activation_perL = (2 * batch_size * sequence_size * hidden_size + 
                               2 * batch_size * sequence_size * rank) * element_size / (10**6)
    param_memory = sum(p.numel() * p.element_size() for p in model.parameters()) / (10**6)
    optimizer_memory = 3 * sum(p.numel() * p.element_size() for p in model.parameters() if p.requires_grad) / (10**6)
    
    return {'static': static_activation_perL, 
            'dynamic': dynamic_activation_perL,
            'param': param_memory,
            'optimizer': optimizer_memory}

# Model Configurations
lora_layer = 12
rank = 16
model = AutoModelForImageClassification.from_pretrained('google/vit-base-patch16-224-in21k')
config = LoraConfig(
    r=rank,
    lora_alpha=16,
    target_modules=["query", "value"],
    lora_dropout=0.1,
    bias="none"
)
net_glob = get_peft_model(model, config)

# Context memory based on experimental values
context_memory_6_layer = 2280  # MB
context_memory_12_layer = 5800  # MB

# Activation Memory Calculation
sequence_size = 197
hidden_size = 768
batch_size = 496
memory_calculator = get_activation(model, rank, sequence_size, hidden_size, batch_size)

# Full 12 Layers
optimizer_memory = memory_calculator['optimizer'] / lora_layer * 12
model_memory = memory_calculator['param']
activation_memory = memory_calculator['static'] * 12 + memory_calculator['dynamic'] * 12
print(f"12 Layers Memory Consumption: {(optimizer_memory + model_memory + activation_memory + context_memory_12_layer) / (10**3)} GB\n")

# First 6 Layers
optimizer_memory = memory_calculator['optimizer'] / lora_layer * 6
model_memory = memory_calculator['param']
activation_memory = memory_calculator['static'] * 12 + memory_calculator['dynamic'] * 6
print(f"First 6 Layers Memory Consumption: {(optimizer_memory + model_memory + activation_memory + context_memory_6_layer) / (10**3)} GB\n")

# Last 6 Layers
optimizer_memory = memory_calculator['optimizer'] / lora_layer * 6
model_memory = memory_calculator['param']
activation_memory = memory_calculator['static'] * 6 + memory_calculator['dynamic'] * 6
print(f"Last 6 Layers Memory Consumption: {(optimizer_memory + model_memory + activation_memory + context_memory_6_layer) / (10**3)} GB\n")
\end{lstlisting}

The estimated memory consumption for different configurations is presented in Table~\ref{tab:memory}.

\begin{table}[t]
    \centering
    \caption{Memory consumption estimates for training different LoRA modules.}
      \scalebox{1.2}{
    \begin{tabular}{c|c}
        \toprule[1pt]
        \textbf{Configuration} & \textbf{Memory Consumption (GB)} \\
        \hline
        Full 12 modules & 47.77 \\
        First 6 modules & 40.57 \\
        Last 6 modules & 23.44 \\
        \bottomrule[1pt]
    \end{tabular}
    }
    \label{tab:memory}
\end{table}

\section{Details on Experimental Settings}\label{sec:appendix-settings}

    \subsection{FL Settings}\label{subsec:appendix-settings-flsettings}

    \begin{table}[t]
      \centering
      \caption{FL settings for all experiments.}
      \scalebox{0.65}{
        \begin{tabular}{c|ccccccccc}
        \toprule[1pt]
        \textbf{Dataset/Configuration} & \textbf{Model} & \textbf{LoRA} & \textbf{Client} & \textbf{Training Round} & \textbf{Learning Rate} & \textbf{batch size} & $T_{\text{IG}}$ & $T_{\text{Agg}}$ & $|D_{\text{IG}}|$ \\
        \midrule
        \textbf{Natural Instruction} & Llama (Data-Juicer-1B) & rank=8, lora\_alpha=16, target\_modules=[``q\_proj'', ``v\_proj''] & 50 (10\%) & 6     & 0.00005 & 8     & 1     & 2     & 50 \\
        \textbf{Dolly-15K} & OPT-1.3B & rank=8, lora\_alpha=16, target\_modules=[``q\_proj'', ``v\_proj''] & 50 (10\%) & 25    & 0.00005 & 8     & 2     & 2     & 50 \\
        \textbf{CIFAR-100} & ViT-base & rank=16, lora\_alpha=16, target\_modules=[``query'', ``value''] & 100 (10\%) & 350   & 0.1   & 496   & 10    & 10    & 50 \\
        \textbf{LEDGAR} & BERT-base & rank=16, lora\_alpha=16, target\_modules=[``query'', ``value''] & 100 (10\%) & 350   & 0.01  & 496   & 10    & 10    & 50 \\
        \textbf{DomainNet-121} & ViT-base & rank=16, lora\_alpha=16, target\_modules=[``query'', ``value''] & 100 (10\%) & 350   & 0.05  & 496   & 10    & 10    & 50 \\
        \bottomrule[1pt]
        \end{tabular}%
        }
      \label{tab:appendix-fl-settings}%
    \end{table}%





    The table presents the FL settings used for all experiments, detailing the datasets, models, LoRA configurations, client distribution, training rounds, learning rates, batch sizes, and key hyperparameters. It includes five datasets: Natural Instruction, Dolly-15K, CIFAR-100, LEDGAR, and DomainNet-121. Different models were used, including Llama (Data-Juicer-1B), OPT-1.3B, ViT-base, and BERT-base. The LoRA configuration varies by dataset, specifying rank, scaling parameter (lora\_alpha), and target modules such as $q_{\text{proj}}$, $v_{\text{proj}}$, $query$, and $value$. The number of participating clients differs, with CIFAR-100, LEDGAR, and DomainNet-121 using 100 clients (10\% per round), while Natural Instruction and Dolly-15K involve 50 clients (10\% per round). Training rounds range from 6 to 350, with learning rates varying between 0.00005 and 0.1. Batch sizes are set to 8 and 496 for different models. 
    Across all experiments, the local IG dataset size $|D_{\text{IG}}|$ is fixed at 50, while other hyper-parameters are case-by-case.

    \subsection{GPU Memory Capacity Settings}\label{subsec:appendix-capacity-settings}

    \begin{table}[t]
      \centering
      \caption{Memory capacity settings and context memory consumptions for heterogeneous client groups.}
      \scalebox{0.85}{
        \begin{tabular}{c|cc|cc|cc|cc}
        \toprule[1pt]
        \multirow{2}[4]{*}{\textbf{Model/Capacity Level}} & \multicolumn{2}{c|}{\textbf{Level 1 (4/10 clients)}} & \multicolumn{2}{c|}{\textbf{Level 2 (3/10)}} & \multicolumn{2}{c|}{\textbf{Level 3 (2/10)}} & \multicolumn{2}{c}{\textbf{Level 4 (1/10)}} \\
    \cmidrule{2-9}          & \textbf{GPU Type} & \textbf{Context (MB)} & \textbf{GPU Type} & \textbf{Context (MB)} & \textbf{GPU Type} & \textbf{Context (MB)} & \textbf{GPU Type} & \multicolumn{1}{c}{\textbf{Context (MB)}} \\
        \midrule
        \textbf{Llama (Data-Juicer-1B)} & V100 32 GB & 80    & A100 40 GB & 1,470  & A6000 48 GB & 2,870  & H100 80 GB & 5,650 \\
        \textbf{OPT-1.3B} & V100 32 GB & 2,890  & A100 40 GB & 4,913  & A6000 48 GB & 6,862  & H100 80 GB & 8,811 \\
        \textbf{ViT-base} & A5000 24 GB & 380   & V100 32 GB & 2,280  & A100 40 GB & 4,170  & A6000 48 GB & 5,800 \\
        \textbf{BERT-base} & A5000 24 GB & 280   & V100 32 GB & 1,962  & A100 40 GB & 3,981  & A6000 48 GB & 5,700 \\
        \bottomrule[1pt]
        \end{tabular}%
        }
      \label{tab:appendix-memory-capacity}%
    \end{table}%

    Table~\ref{tab:appendix-memory-capacity} categorizes memory capacity settings and GPU Context memory consumption for different models across four client levels based on experimental results. Llama (Data-Juicer-1.3B) has the lowest context memory at 80 MB on Level 1 (V100 32 GB) and the highest at 5650 MB on Level 4 (H100 80 GB). OPT-1.3B follows a similar trend, ranging from 2890 MB on Level 1 to 8811 MB on Level 4. ViT-base starts at 380 MB on Level 1 (A5000 24 GB) and increases to 5800 MB on Level 4 (A6000 48 GB). BERT-base consumes the least memory, from 280 MB on Level 1 to 5700 MB on Level 4. These results illustrate the scaling of context memory with GPU capacity and the computational demands of different models under heterogeneous client conditions.

    \uline{The GPU configurations are flexible, and our Knapsack optimization algorithm can be applied to various GPU devices with various memory capacities.}

    \subsection{Experimental Settings for Figure~\ref{fig:motivation}}\label{subsec:appendix-motivation-settings}

    In Figure~\ref{fig:motivation}, we present preliminary experiments to explore how different LoRA allocation strategies impact memory consumption and global model accuracy. The experiments use ViT-base, a transformer-based model with 12 encoder layers, where LoRA is applied to all 12 layers, specifically to the “query” and “value” components. The hyperparameters are set as follows: rank = 16, alpha = 16, and dropout ratio = 0.1. In the FL settings, the total number of users is 100, the selection rate is 10\%, the total training rounds are 350, and the local round is set to 1. The batch size is 496, distributed across 8 GPUs (62 on each GPU). The local learning rate is 0.01, optimized using Adam. Results are evaluated on CIFAR-100 and divided according to an IID distribution.


    we conduct experiments by training the first 1, first 6, last 1, last 6, and all 12 LoRA modules on the CIFAR-100 dataset, as shown in Figure.~\ref{fig:motivation} (a). These strategies lead to varying memory consumption, which corresponds to different memory capacities commonly seen in real-world NVIDIA GPUs, such as NVIDIA RTX A2000 (12GB), NVIDIA RTX A5000 (24GB), NVIDIA Tesla A100 (40GB), NVIDIA RTX A6000 (48GB). The GPU memory usage is calculated based on the method described in Section~\ref{subsec:appendix-memory-footprint-example}. The memory consumption for contexts with 1, 6, and 12 layers is derived from multiple experimental results and fixed at 100, 2,280, and 5,800 MB, respectively.

    Due to the presence of static activations~\cite{ardakani2024slimfit}, which cannot be discarded because of non-linear functions (e.g., GELU, LayerNorm, and Softmax), the memory consumption for training the first 1 or 6 layers approaches that of training all 12 layers. Additionally, the results reveal a notable performance disparity based on which LoRA modules are trained (e.g., a comparison between the ``Last 6'' and ``First 6'' strategies), highlighting the varying importance of different layers in the model.
    
\section{Details of Dataset}\label{sec:appendix-dataset}

\circled{1} \textbf{Natural Instruction:}
The Natural Instruction dataset~\cite{wang2022super} is a large-scale collection of human-written instructions designed to facilitate instruction-based learning for natural language processing models. It consists of a diverse set of tasks, covering various domains such as classification, summarization, translation, question answering, and reasoning. The dataset consists of 1,616 diverse natural language tasks with instructions and answers. 
In our experiments, data instances are formatted into prompts before being processed by the LLMs. We utilize the template provided by Alpaca~\cite{taori2023stanford} to format the datasets used in our experiments.
In the pre-processing, we first filter out tokens with a total length larger than 512, which results in 613 tasks, with 50,348 training samples, and 5,517 testing samples. 
%
We simulate three levels of task heterogeneity from IID to extreme Non-IID, by varying the number of client's data tasks to full, 20, and 15 for NI.

\circled{2} \textbf{Dolly-15K:} The Dolly-15K dataset~\cite{conover2023free} is an open-source instruction-tuning dataset introduced by Databricks to improve the fine-tuning of large language models for instruction-following tasks. It comprises 15,000 human-generated prompt-response pairs spanning diverse domains such as open-ended question answering, summarization, brainstorming, and information extraction. Unlike many instruction-tuning datasets that rely on synthetic data, Dolly-15K ensures high-quality, natural language interactions through human-authored responses, making it a valuable resource for training and evaluating instruction-following models.
For preprocessing, we applied the same prompts as in the NI dataset and filtered out samples exceeding 512 tokens in total length. This resulted in 10,936 training and 2,735 testing samples. To simulate different levels of task heterogeneity, we varied the number of tasks per client, creating three settings: IID (full task set per client), mild Non-IID (3 tasks per client), and extreme Non-IID (1 task per client).

\circled{3} \textbf{CIFAR-100:} The CIFAR-100 dataset consists of 60,000 32$\times$32 color images in 100 classes, with 600 images per class. There are 500 training images and 100 testing images per class. For the Non-IID data distribution, we apply quantity and label skews to the dataset. For quantity skew, we use Dirichlet with $\alpha=1.0$, and for label skew, we use pathological where the data on each client only contains the specific number of labels. This results in three levels of data distribution: $\{IID, (20/Dir\sim1.0), (10/Dir\sim1.0)\}$, where $(20/Dir\sim1.0)$ means each client only contains 20 out of 100 categories of samples, and the number of samples obeys the Dirichlet distribution with $\alpha=1.0$. 

\circled{4} \textbf{LEDGAR-100:} The LEDGAR dataset aims at contract provision classification. The contract provisions sourced from contracts obtained from the US Securities and Exchange Commission filings, which are publicly available from EDGAR. Each category represents the single main topic of the corresponding contract provision. There are 100 categories, with 60,000 samples for training and 10,000 for testing. The number of samples for each category is unbalanced. The data distribution skew is the same as that of CIFAR-100. 

\circled{5} \textbf{DomainNet-121:} The original DomainNet is a dataset of common objects in six different domains. All domains include 345 categories of objects and a total of {424.5k} images. The domains include clipart, real, sketch, infograph, painting, and quickdraw. We craft DomainNet-121 by selecting categories that all six domains have more than 50 images, resulting in 121 categories. Then, we randomly choose 50 images per category per domain, resulting in 300 images per category. Subsequently, we sample 5 images from each category in each domain for testing. There are 270 training images and 30 testing images per class. For Non-IID data distribution, we apply domain and category skews: $\{1D/121C, 1D/50C, 1D/25C)\}$, where $1D/121$ means each client contains 121 out of 121 categories and 1 out of 6 domains of samples.  

More specifically, the categories drawn from the original DomainNet are as follows:

\begin{multicols}{4}
\noindent
barn\\
baseball\_bat\\
basket\\
beach\\
bear\\
beard\\
bee\\
bird\\
blueberry\\
bowtie\\
bracelet\\
brain\\
bread\\
broccoli\\
bus\\
butterfly\\
circle\\
cloud\\
cruise\_ship\\
dolphin\\
dumbbell\\
elephant\\
eye\\
eyeglasses\\
feather\\
fish\\
flower\\
foot\\
frog\\
giraffe\\
goatee\\
golf\_club\\
grapes\\
grass\\
guitar\\
hamburger\\
hand\\
hat\\
headphones\\
helicopter\\
hexagon\\
hockey\_stick\\
horse\\
hourglass\\
house\\
ice\_cream\\
jacket\\
ladder\\
leg\\
lipstick\\
megaphone\\
monkey\\
moon\\
mushroom\\
necklace\\
owl\\
panda\\
pear\\
peas\\
penguin\\
pig\\
pillow\\
pineapple\\
pizza\\
pool\\
popsicle\\
rabbit\\
rhinoceros\\
rifle\\
river\\
sailboat\\
sandwich\\
sea\_turtle\\
shark\\
shoe\\
skyscraper\\
snorkel\\
snowman\\
soccer\_ball\\
speedboat\\
spider\\
spoon\\
square\\
squirrel\\
stethoscope\\
strawberry\\
streetlight\\
submarine\\
suitcase\\
sun\\
sweater\\
sword\\
table\\
teapot\\
teddy-bear\\
telephone\\
tent\\
The\_Eiffel\_Tower\\
The\_Great\_Wall\_of\_China\\
The\_Mona\_Lisa\\
tiger\\
toaster\\
tooth\\
tornado\\
tractor\\
train\\
tree\\
triangle\\
trombone\\
truck\\
trumpet\\
umbrella\\
vase\\
violin\\
watermelon\\
whale\\
windmill\\
wine\_glass\\
yoga\\
zebra\\
zigzag
\end{multicols}

\textbf{Metric for each dataset.} For NI and Dolly-15K datasets, we use the Rouge-L metric. Rouge-L (Recall-Oriented Understudy for Gisting Evaluation - Longest Common Subsequence) is a widely used metric for evaluating the quality of text generated by large language models, particularly in summarization and text generation tasks. Unlike Rouge-N, which measures exact n-gram overlap, Rouge-L focuses on the Longest Common Subsequence (LCS) between the generated text and the reference text, capturing sentence-level fluency and coherence. Since LCS does not require consecutive word matches, it is better suited for evaluating paraphrased or loosely structured responses. The Rouge-L score consists of precision, recall, and F1-score, providing a balanced measure of how well the generated text aligns with the reference while allowing for flexible word order. This makes it particularly useful for assessing LLM outputs in real-world applications where exact phrasing may vary while preserving meaning. For CIFAR-100 and DomainNet-121 datasets, we evaluate the performance using accuracy, defined as the ratio of the number of correct predictions to the total number of predictions, i.e., $\text{Accuracy} = \frac{\text{Number of Correct Predictions}}{\text{Total Number of Predictions}} \times 100\%$.
In the case of the LEDGAR dataset, we compute the metrics Micro F1 and Macro F1. Micro F1 is calculated as: $\frac{2 \times \text{Micro Precision} \times \text{Micro Recall}}{\text{Micro Precision} + \text{Micro Recall}}$, where Micro Precision and Micro Recall are defined as $\frac{\sum_{i=1}^{N} TP_i}{\sum_{i=1}^{N} (TP_i + FP_i)}$ and $\frac{\sum_{i=1}^{N} TP_i}{\sum_{i=1}^{N} (TP_i + FN_i)}$, respectively. Here, $TP_i$ is the number of true positive predictions for class $i$, $FP_i$ is the number of false positive predictions for class $i$, $N$ is the total number of classes. On the other hand, Macro F1 is calculated individually for each class. It is computed by replacing Micro Precision and Micro Recall with Macro Precision and Macro Recall which are defined as, $\frac{1}{N} \sum_{i=1}^{N} \frac{TP_i}{TP_i + FP_i}$ and $\frac{1}{N} \sum_{i=1}^{N} \frac{TP_i}{TP_i + FN_i}$, respectively. We use the Macro F1 score to evaluate the best result for the LEDGAR dataset.

\section{More Results}\label{sec:appendix-results-visualization}

\begin{figure}[t]
\centering
\includegraphics[width=0.45\textwidth]{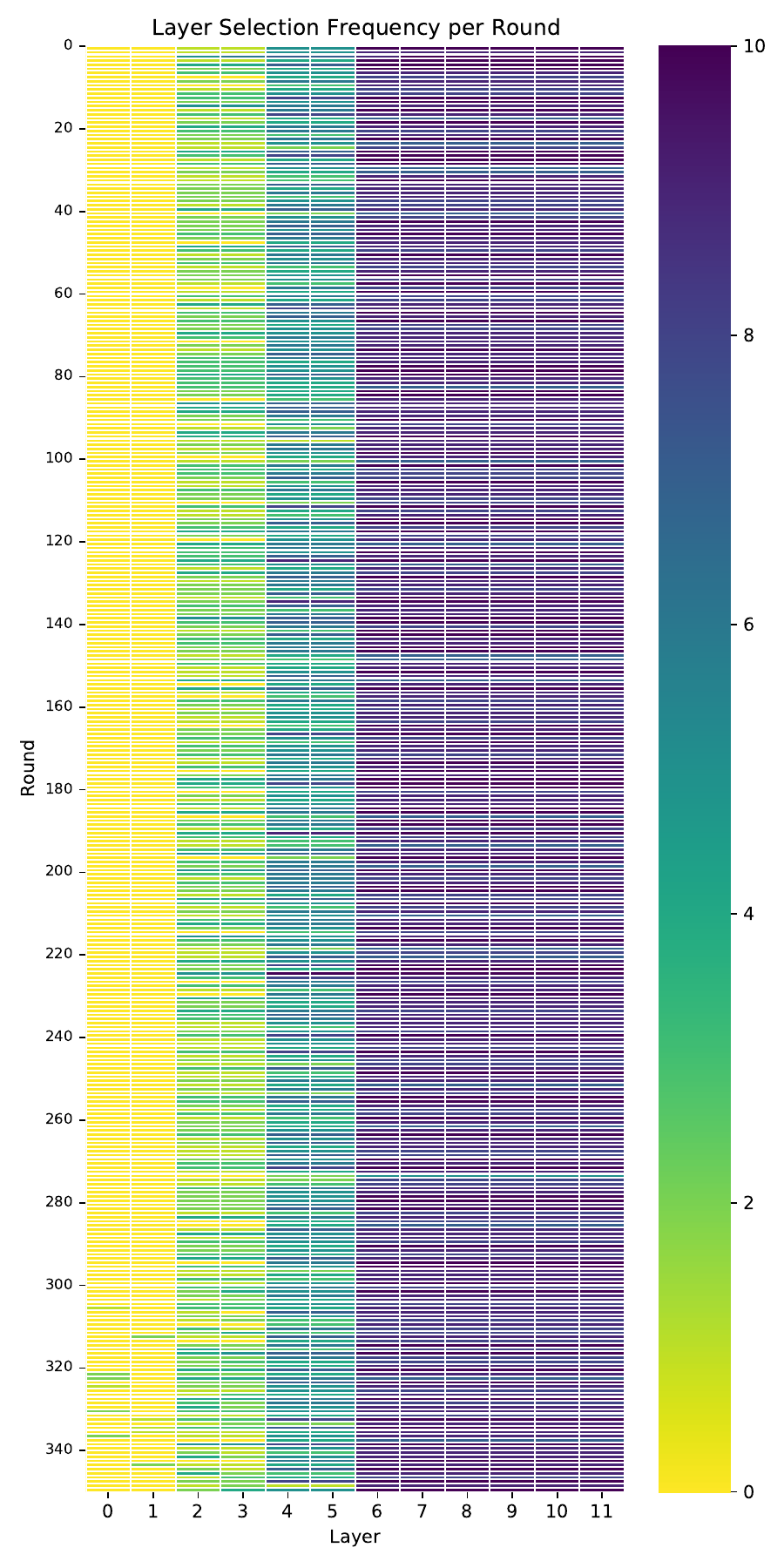}
\caption{The selection frequency of LoRA modules in each attention layer across rounds.}
\label{fig:layerselection}
\end{figure}

\textbf{The LoRA Module Allocation Frequency.} We visualize the selection frequency of LoRA modules in each attention layer across training rounds. The experiment is conducted on the CIFAR-100 dataset under the Non-IID (10) setting. As shown in Figure~\ref{fig:layerselection}, our allocation optimization algorithm demonstrates stable LoRA selection patterns across rounds. The last six layers (layers 6 to 11) are most frequently selected throughout training. Layers 4 and 5 are chosen more often than layers 2 and 3 in most rounds. The first two layers are selected less frequently due to their higher memory consumption but are occasionally chosen in later rounds as other layers converge.

\section{Comparative Study on Related Works}\label{sec:appendix-comparative_study}

\begin{table}[t]
  \centering
  \caption{A comparative study with state-of-the-art works.}
  \scalebox{0.90}{
    \begin{tabular}{ccccccccc}
    \toprule[1pt]
    \textbf{Method}   & \textbf{PEFT} & \textbf{Resource Heterogeneity} & \textbf{Parameter Heterogeneity} & \textbf{Dynamic Training}  \\\hline
    {FedIT~\cite{zhang2023towards}}               & {\ding{51}}        & {\ding{55}}        & {\ding{55}}         & {\ding{55}}                   \\
    {SLoRA~\cite{babakniya2023slora}}                      & {\ding{51}}        & {\ding{55}}        & {\ding{55}}        &  {\ding{55}}                \\
    {FedMS~\cite{wu2024fedfmsl}}                      &{\ding{51}}         & {\ding{55}}        & {\ding{55}}        &  {\ding{51}}                 \\
    {FwdLLM~\cite{xu2024fwdllm}}                     & {\ding{51}}        & {\ding{55}}        & {\ding{55}}        & {\ding{55}}                  \\
    {FFA-LoRA~\cite{sun2024improving}}                      & {\ding{51}}        & {\ding{55}}        & {\ding{55}}        & {\ding{55}}                  \\
    HETLoRA~\cite{cho2024heterogeneous}                   & {\ding{51}}        & {\ding{51}}        & {\ding{51}}        &  {\ding{55}}              \\
    {FLoRA~\cite{wangflora}}                          & {\ding{51}}        & {\ding{51}}        & {\ding{51}}        & {\ding{55}}               \\
    {FlexLoRA~\cite{bai2024federated}}                         & {\ding{51}}        & {\ding{51}}        & {\ding{51}}        & {\ding{55}}               \\
    {FedRA~\cite{su2023fedra}}                          &{\ding{51}}         & {\ding{51}}        & {\ding{51}}        & {\ding{55}}                 \\
    \midrule
    \textbf{{Ours}}                   & {\ding{51}} & {\ding{51}} & {\ding{51}} & {\ding{51}}  \\
    \bottomrule[1pt]
    \end{tabular}%
    }
  \label{tab:survey}%
\end{table}%

In this section, we compare our approach with existing methods that are related to ours in terms of technical aspects and objectives.
%
%
%
In Table~\ref{tab:survey}, we present a comparative analysis of our method against state-of-the-art approaches in federated fine-tuning. The comparison considers four key aspects: whether is a PEFT (Parameter-Efficient Fine-Tuning)-based method, whether considers resource heterogeneity, whether considers trainable parameter heterogeneity, and whether incorporate dynamic training. While all listed methods support PEFT, most do not address resource heterogeneity, limiting their adaptability to devices with varying memory capacities. Methods such as HETLoRA, FLoRA, and FlexLoRA incorporate trainable parameter heterogeneity, allowing clients to train LoRA modules with different ranks, but they fail to accommodate resource constraints effectively. Additionally, only a few methods, including FedMS and FedBiOT, support dynamic training, where models can adapt their configurations during training. Our proposed method is the only approach that simultaneously supports PEFT, resource heterogeneity, trainable parameter heterogeneity, and dynamic training, making it a more flexible and scalable solution for federated fine-tuning in heterogeneous environments.

\section{Limitations}\label{sec:appendix-limitation}

Our LoRA allocation optimization is highly adaptable to various memory capacity levels, supporting devices ranging from consumer-grade GPUs to datacenter-grade GPUs. However, if a client cannot train even a single LoRA module, our method can easily combined with other techniques, such as offloading~\cite{kar2023offloading}, checkpointing~\cite{rasley2020deepspeed} to ensure participation without exceeding hardware limitations.

\section{Future Work}\label{sec:appendix-futurework}

{Our method is orthogonal to aggregation noise-reduction techniques} such as FlexLoRA~\cite{bai2024federated} and FLoRA~\cite{wangflora}, allowing for seamless integration to heterogeneous model aggregation. In future work, we plan to incorporate these methods while preserving optimal LoRA allocation across heterogeneous devices.

\end{document}